\documentclass[10pt,twocolumn,letterpaper]{article}

\usepackage[pagenumbers]{cvpr} 

\usepackage{graphicx}
\usepackage{amsmath}
\usepackage{amssymb}
\usepackage{booktabs}

\usepackage[pagebackref,breaklinks,colorlinks]{hyperref}
\usepackage{import}
\usepackage{easy-todo}
\usepackage{makecell}
\usepackage{xcolor}
\usepackage{enumitem}

\usepackage[capitalize]{cleveref}
\crefname{section}{Sec.}{Secs.}
\Crefname{section}{Section}{Sections}
\Crefname{table}{Table}{Tables}
\crefname{table}{Tab.}{Tabs.}

\def\cvprPaperID{6748} 
\def\confName{CVPR}
\def\confYear{2023}

\begin{document}

\title{Enhancing Targeted Attack Transferability via Diversified Weight Pruning}

\author{Hung-Jui Wang, Yu-Yu Wu, Shang-Tse Chen\\
National Taiwan University\\
{\tt\small \{r10922061, r10922018, stchen\}@csie.ntu.edu.tw}
}
\maketitle

\newcommand{\newborn}{pruned}
\newcommand{\baseline}{NI-SI-TI-DI}
\newcommand{\ghost}{GN}
\newcommand{\dual}{DSNE}
\newcommand{\demoImgWidth}{0.16\linewidth}

\begin{abstract}
Malicious attackers can generate targeted adversarial examples by imposing tiny noises, forcing neural networks to produce specific incorrect outputs. With cross-model transferability, network models remain vulnerable even in black-box settings. Recent studies have shown the effectiveness of ensemble-based methods in generating transferable adversarial examples. To further enhance transferability, model augmentation methods aim to produce more networks participating in the ensemble. However, existing model augmentation methods are only proven effective in untargeted attacks. In this work, we propose Diversified Weight Pruning (DWP), a novel model augmentation technique for generating transferable targeted attacks. DWP leverages the weight pruning method commonly used in model compression. Compared with prior work, DWP protects necessary connections and ensures the diversity of the pruned models simultaneously, which we show are crucial for targeted transferability. Experiments on the ImageNet-compatible dataset under various and more challenging scenarios confirm the effectiveness: transferring to adversarially trained models, Non-CNN architectures, and Google Cloud Vision. The results show that our proposed DWP improves the targeted attack success rates with up to $10.1$\%, $6.6$\%, and $7.0$\% on the combination of state-of-the-art methods, respectively. The source code will be made available after acceptance.
\end{abstract}

\section{Introduction}

\begin{figure}[ht]
\centering
\includegraphics[width=\columnwidth]{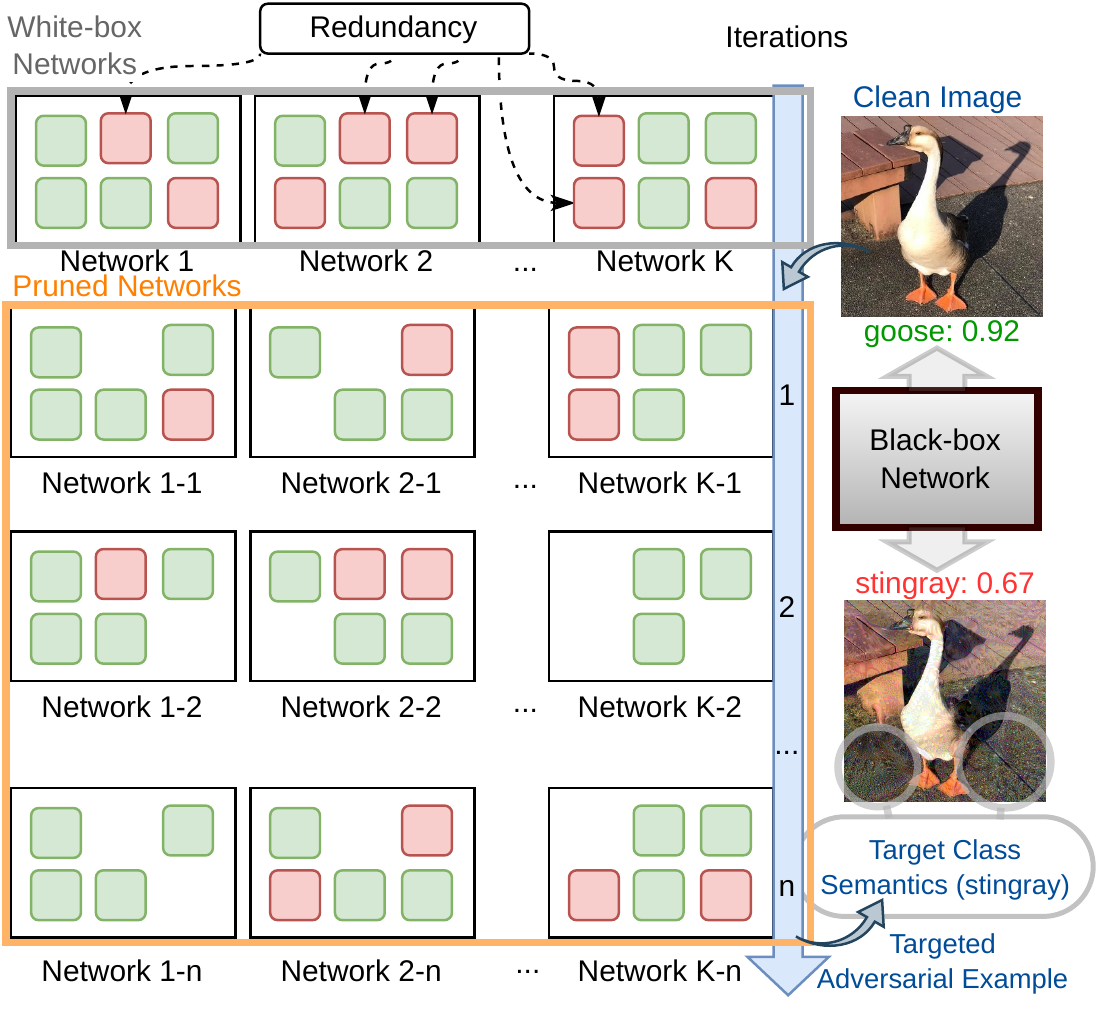}
\label{fig:intro_img}
\vspace{-5mm}
\caption{\textbf{The big picture of DWP.} Based on the over-parameterized property of neural networks, we leverage weight pruning to produce additional diversified \newborn{} models from existing white-box networks at each iteration. By protecting necessary weight connections in each network, the quality of models is well-preserved. These additional pruned models can better impose the semantics of the target class onto adversarial images, yielding higher targeted transferability.}
\end{figure}

While deep learning continues to achieve breakthroughs in various domains, recent studies have shown vulnerabilities of deep neural networks to adversarial attacks, causing severe threats in safety-critical applications. For example, in image classification, an attacker can add human-imperceptible perturbations to benign images at testing time. These adversarial examples can fool a well-trained neural network to yield arbitrary classification results. Several attacks have been proposed to improve and evaluate the robustness of CNNs~\cite{IEEESP_CW, ICLR_ens_AdvTraining, NIPS_Cross}.

In the white-box settings, with complete information on the victim model, the attacker can generate adversarial examples effectively and efficiently.  As for black-box settings, where the attacker only has limited information about the victim model, it is still possible to create cross-model attacks using a substitute model with white-box adversarial attack methods. This kind of black-box attack depends on the transferability of adversarial attacks.

Many methods have been proposed to increase the transferability for untargeted attacks, where the goal is to decrease the accuracy of the victim model. However, there is still room for improvement in creating transferable targeted attacks, where the attacker aims to mislead the victim model to produce a predefined specific outcome. Recent works use an ensemble-based approach to generate transferable targeted adversarial examples with multiple neural networks as substitute models simultaneously~\cite{DelvingLiu2017,NIPS_Simple}. To further enlarge the power of the ensemble, model augmentation creates additional networks by altering the existing ones \cite{AAAI_Ghost, dual_erosion} and generates adversarial examples with these networks altogether.

However, we find that current methods of model augmentation rarely consider the importance of neurons and weight connections in networks.
While Ghost Networks \cite{AAAI_Ghost} inserts extra dropout layers and random skip connection mechanisms into networks to produce additional models, these dropout layers randomly drop neurons away without considering their significance. Authors in \cite{dual_erosion} introduce another uniform erosion on the remaining parameters after dropout and skip connection to increase diversity. However, there is still a lack of protection on necessary parameters. To avoid excessively destroying the performance of networks, these methods require heavy tuning on the hyperparameters like dropout rates, the amount of skip connection, the second erosion rates, and the locations of the inserted dropout layers. 

When it comes to transferable targeted attacks, the quality of white-box substitute models plays a more crucial role. Rather than merely moving away from the original class, the semantics of targeted adversarial examples need to be close to the target class to acquire higher transferability \cite{CVPR_Poincare, ICCV2021_GAN}. Dropping or disturbing the significant components in substitute networks can mislead targeted adversarial examples and yield worse transferability.

To overcome these problems, we learn from model compression and propose an improved model augmentation method named Diversified Weight Pruning (DWP). Model compression reduces the storage and computation overhead without substantial influence on model performances \cite{NIPS_OptimalBrain, NIPS_L1Pruning, ICLR_Lottery, ICLR_RethinkingPruning}. With the over-parameterized property \cite{NIPS_overparameterized} of neural networks, weight pruning \cite{NIPS_L1Pruning} can maintain the performance of a network by only removing redundant weight connections. To generate transferable targeted adversarial examples, we apply random weight pruning to each single CNN network accessible to form additional ones. These pruned networks remain stable since the significant weight connections are protected. We thus improve the ensemble-based approach with these extra diverse models.

To evaluate DWP, we experiment with an ImageNet-compatible dataset used in the NIPS 2017 adversarial attack competition \cite{NIPS2017_competition}. The average targeted success rate of DWP reaches $81.30$\% across CNNs. Furthermore, we test DWP in the more challenging scenarios of transferring to adversarially trained models and Non-CNN architectures. The results show that DWP improves the targeted success rate with up to $10.1$\% and $6.6$\% on average in these two settings. Finally, we demonstrate our DWP on the real-world Google Cloud Vision service and get $7.0$\% improvement.

In summary, our primary contributions are as follows:
\begin{itemize}[itemsep=0em,topsep=1pt]
    \item We propose DWP leveraging weight pruning to improve the existing model augmentation methods on transferable targeted attacks. Experiments show that our DWP enhances the combination of current state-of-the-art techniques.
    \item The experiment results show that DWP remains effective in more challenging settings like transferring to adversarially trained models, Non-CNN architectures, and even the real-world Google Cloud Vision service.
    \item We analyze the cosine similarities of adversarial perturbations between different pruned networks to verify that DWP increases the diversity of networks for generating adversarial perturbations.
    \item We provide intuitive experiments on explaining the success of targeted attacks with DWP.
\end{itemize}
\section{Related Work}
\subsection{Transferable Attack}
Throughout the work, we focus on simple transferable attacks \cite{NIPS_Simple}, which require neither additional data nor model training for attacking compared to resource-intensive attacks. Recent works aiming for simple transferable attacks mainly include four categories: gradient optimization, input transformation, advanced loss function, ensemble, and model augmentation.
\subsubsection{Gradient Optimization}
Optimization-based methods are widely adopted \cite{IntriguingPropertiesAdv, HarnessAdv, IEEESP_CW, ICLR_BIM} in generating adversarial examples. With iterative methods \cite{ICLR_BIM, IEEESP_CW}, one can get better solutions to an objective function for attacking through multiple times of optimization on adversarial examples and get stronger attacking results. Adjusting gradients used to update adversarial examples at each iteration appropriately has been shown beneficial for overcoming sub-optimal results in optimization.
\cite{BoostingMomentum} combines momentum techniques with iterative attacks, accumulating gradients at each iteration to escape local optimum and stable the direction of updating.
\cite{ICLR_Nesterov} applies Nesterov accelerated gradient for optimization, giving adversarial examples an anticipatory updating to yield faster convergence.
\subsubsection{Input Transformation}
Motivated by Data Augmentation \cite{BigData_DataAugmentation}, several works suggest attacking transformed input to prevent adversarial examples from overfitting white-box models and failing to transfer to black-box ones.
\cite{CVPRDI} uses random resizing and padding throughout the iterative attack. \cite{CVPRTI} enumerates several translated versions for each input image and fuses the gradients acquired on all of them. \cite{ICLR_Nesterov} leverages the scale-invariant property of CNNs and employs multiple scale copies from each input image. \cite{ICCV_Admix} extends the concept of mixup \cite{ICLR_Mixup}, attacking the mixup version of each input image.

\subsubsection{Modern Loss Function}
Cross entropy loss is widely used in image classification, also serving as the objective function for adversarial attacks. However, for targeted attacks, cross entropy is pointed out the saturation problem \cite{CVPR_Poincare} as the output confidence of target class approaches to one. To this end, alternative loss functions attempt to provide more suitable gradients for optimization. \cite{CVPR_Poincare} leverages Poincar\'e distance as the loss function, which amplifies the gradient magnitude as the confidence of the target class grows. \cite{NIPS_Simple} proposes a simple logit loss, which has constant gradient magnitude regardless of the output probability.

\subsubsection{Ensemble and Model Augmentation}
Adversarial examples generated by ensembling multiple white-box networks are more likely to transfer to black-box networks \cite{DelvingLiu2017}. Instead of simply fusing the output confidence of each white-box network, \cite{CVPR_red_var_ensemble} suggests reducing the gradient variance of white-box models during attacking.
To further improve ensemble-based approaches, Model Augmentation produces additional diverse models from the existing white-box networks. \cite{AutoMA} uses reinforcement learning to automatically find transformations suitable with white-box networks to yield more diversity. \cite{AAAI_Ghost} acquires ghost networks for ensemble through perturbing dropout and skip connections of existing ones. \cite{dual_erosion} further improves the diversified ensemble via dual-stage erosion.

\subsection{Network Pruning}
The intensive cost of computation and storage hinders applications of neural networks, especially on embedding systems. Network Compression aims to reduce the scale of networks, making them more feasible for deployment. With the over-parameterized property \cite{NIPS_overparameterized}, several works about removing redundancy in networks, known as Network Pruning, are proposed and become a branch of Network Compression. \cite{NIPS_OptimalBrain} uses the second-derivative information to find redundant weights in networks. \cite{NIPS_L1Pruning} shows that neural networks can highly preserve performance even if trimming more than half of their connections. Retraining after pruning for better preservation of accuracy is also investigated \cite{ICLR_Lottery, ICLR_RethinkingPruning}.
\section{Methodology}
Given a neural network $\theta$ and a benign example $x$, we generate a targeted adversarial example $x^{\textrm{adv}}$ for the target class $y^{\textrm{target}}$ by solving the following constrained optimization problem:
$$\mathop{\arg\min}_{x^{\textrm{adv}}} J(x^{\textrm{adv}}, y^{\textrm{target}};\theta) \quad \textrm{s.t.} \quad \left\|x^{\textrm{adv}}-x\right\|_\infty \leq \epsilon,$$
where $J$ is the loss function for multiclass classification and $\epsilon$ is the perturbation budget. To circumvent the gradient saturation problem of cross-entropy, we use logit loss~\cite{NIPS_Simple} as our loss function $J$.
\subsection{Preliminary and Motivation}
We start by establishing the roles of current state-of-the-art techniques in our iterative attack. Then, we demonstrate how we apply Weight Pruning to improve targeted transferability.

\subsubsection{Momentum and Nesterov Iterative Method (NI) \cite{BoostingMomentum, ICLR_Nesterov}}

Inspired by Nesterov Accelerated Gradient \cite{1983nesterov}, Nesterov Iterative Method (NI) modifies Momentum Iterative-FGSM \cite{BoostingMomentum} by adding the historical gradients to current adversarial examples $x_n$ and gets $x^{\textrm{nes}}_n$ in advance. Gradients at the ahead $x^{\textrm{nes}}_n$ instead of the current $x_n$ will be used for updating. The scheme helps accelerate convergence by avoiding the local optimum earlier:
$$x^{\textrm{nes}}_n=x_n+\alpha \cdot \mu \cdot g_{n-1}$$
$$g_n=\mu\cdot g_{n-1}+\nabla_xJ(x^{\textrm{nes}}_n, y^{\text{target}};\theta)$$
$$x_{n+1} = \textrm{Clip}_x^{\epsilon}(x_n - \alpha\cdot\textrm{sign}(g_n)).$$
Here $\mu$ is the decay factor of the historical gradients. The gradient computed encourages adversarial examples to increase confidence logit output by the white-box network model $\theta$ on the $\text{target}$ class through gradient ascent with learning rate $\alpha$. A clipping operation onto the $\epsilon$-ball centered at the original input image $x$ is at the end of each iteration. To preserve more information about the gradient for attacking \cite{AAAI_Indistinguishable}, we don't include the L1 normalization.

\subsubsection{Scale Invariant Method (SI) \cite{ICLR_Nesterov}}
Neural networks can preserve output even though the input image $x$ goes through scale operations such as $S_m(x)=x/2^m$. With the scale-invariant property, each composite of white-box networks and scale operations becomes different functions. Adversarial examples can enjoy more diverse gradients:
$$g_n=\mu\cdot g_{n-1}+\dfrac{1}{M}\sum_{m=0}^{M-1}\nabla_xJ(S_m(x^{\textrm{nes}}_n), y^{\text{target}};\theta).$$
$M$ is the number of scaled versions feeding into the network for each image.

\subsubsection{Diverse Input Patterns (DI) \cite{CVPRDI}}
Inspired by data augmentation techniques \cite{BigData_DataAugmentation} used in network training, DI imposes random resizing and padding on each image before it feeds into network models to avoid overfitting. Straightforward cooperation with NI and SI is as follows:
$$g_n=\mu\cdot g_{n-1}+\dfrac{1}{M}\sum_{m=0}^{M-1}\nabla_xJ(S_m(T(x^{\textrm{nes}}_n, p_{\textrm{DI}})), y^{\text{target}};\theta).$$
The introduced $T$ decides whether to apply random resizing at each iteration with probability $p_{\textrm{DI}}$, which degenerates when $p_{\textrm{DI}}=0$.

\subsubsection{Translation Invariant Method (TI) \cite{CVPRTI}}
To deal with different discriminative regions \cite{CVPRTI} of various defense neural networks, TI produces several translated versions for the current image in advance and computes the gradient for each separately. These gradients will then be fused and used to attack the current image. \cite{CVPRTI} also shows that one can approximate the gradient fusion using convolution. The approximation prevents TI from enduring the costly computation on excessive translated versions for every single image, also yielding the further revised updating procedure:
$$g_n=\mu\cdot g_{n-1}+ \textit{\textbf{W}}* \dfrac{1}{M}\sum_{m=0}^{M-1}\nabla_xJ(S_m(T(x^{\textrm{nes}}_n, p_{\textrm{DI}})), y^{\text{target}};\theta).$$
$\textit{\textbf{W}}$ is the convolution kernel matrix applied. Some typical options are linear, uniform, or Gaussian kernel.

We abbreviate the attacking procedure so far to NI-SI-TI-DI with the combination of these techniques.

\subsection{Diversified Weight Pruning}
We name the proposed approach Diversified Weight Pruning (DWP) due to the increased diversity of white-box models for ensemble via Weight Pruning. Following Weight Pruning, we sort the connections of each white-box network by the L1 norm of their weight values. With a predefined rate $r$, we only consider the lowest ($100\cdot r$)\% ``prunable'' since weights with small absolute values are shown unnecessary \cite{NIPS_L1Pruning}. Networks can preserve accuracy after these connections are pruned away even without retraining \cite{NIPS_L1Pruning}.

For our pruning operation, we first identify the set of prunable weights. Let $\gamma$ be the ($100\cdot(1-r)$)-th percentile of weights in $\theta$. We formulate the prunable set:
$$\Gamma(\theta, r)=\{w\in\theta|w<\gamma\} \subseteq \theta.$$
With $\Gamma(\theta, r)$ collecting all the prunable weights of $\theta$, we introduce an indicator vector for it:
$$\Pi_{\Gamma(\theta, r)}=(\lambda_1, \lambda_2, ..., \lambda_{\kappa}),$$
where $\kappa$ is the total number of weights in $\theta=\{w_1, w_2, ..., w_\kappa\}$ including non-prunable ones. $\lambda_i$ is determined by whether its corresponding $w_i\in\theta$ is in the prunable subset $\Gamma(\theta, r)$:
\[
    \lambda_i= 
    \begin{cases}
        1,      & \text{if } w_i\in \Gamma(\theta, r)\\
        0,      & \text{otherwise}
    \end{cases}.
\]
Supported by the indicator vector $\Pi_{\Gamma(\theta, r)}$, our pruning operation $P(\cdot)$ can protect the non-prunable weights by masking:
$$P(\theta, r)=(\textbf{1}_{\kappa}-\Pi_{\Gamma(\theta, r)}\odot \textbf{b})\odot \theta,$$
where $\odot$ denotes the element-wise multiplication and $\textbf{1}_{\kappa}=(1,1, ..., 1)\in R^\kappa$ denotes an all-one vector. $\textbf{b}=(b_1, b_2, ..., b_\kappa)$ is a vector with $b_i\overset{\textrm{i.i.d}}{\sim}\textrm{Bernoulli}(p_{\textrm{bern}})$, where $p_{\textrm{bern}}$
is the probability for pruning each connection independently.

To be specific, $\Pi_{\Gamma(\theta, r)}$ and $\textbf{b}$ both are binary masks with identical layout as $\theta$. $\Pi_{\Gamma(\theta, r)}$ is responsible for protecting non-prunable weights, while $\textbf{b}$ is for random pruning. Each binary element in $\Pi_{\Gamma(\theta, r)}\odot \textbf{b}$ indicates whether to prune the corresponding weight value in $\theta$. The main difference from Dropout \cite{JMLR_Dropout} used in previous model augmentations \cite{AAAI_Ghost, dual_erosion}, is that DWP only considers prunable weights. The detailed comparison will be shown in the following sections.

Instead of producing all the pruned models beforehand, we acquire pruned models at each iteration right before gradient computing.
\[
    \begin{aligned}
        &g_n=\mu\cdot g_{n-1}+\\
        &\dfrac{\textit{\textbf{W}}}{M}* \sum_{m=0}^{M-1}\nabla_xJ(S_m(T(x^{\textrm{nes}}_n, p_{\textrm{DI}})), y^{\text{target}};P(\theta, r)).
    \end{aligned}
\]

With this longitudinal ensemble strategy~\cite{AAAI_Ghost, dual_erosion},
the storage and memory overhead are almost identical to the original attack procedure.
With $K$ white-box models, our final DWP attack procedure is shown as follows:
\[
    \begin{aligned}
        &g_n=\mu\cdot g_{n-1}+\\
        &\dfrac{\textit{\textbf{W}}}{M}* \sum_{m=0}^{M-1}\sum_{k=1}^{K}\beta_{k}\nabla_xJ(S_m(T(x^{\textrm{nes}}_n, p_{\textrm{DI}})), y^{\text{target}};P(\theta_{k}, r)),
    \end{aligned}
\]
where $\beta_k$ are the ensemble weights, $\sum_{k=1}^K\beta_k = 1$.

Benefiting from no dependency on network retraining and extra data, our proposed DWP is simple and lightweight. As there is no further retraining, we select the L1 norm for pruning since it is better than L2 on preserving accuracy \cite{NIPS_L1Pruning}.
\section{Experiments}
In this section, we first describe the experiment settings and demonstrate the results of transferable targeted attacks under various scenarios. We then inspect the diversified property of pruned models in DWP from the view of geometry. Finally, we illustrate an intuitive explanation of the success of transferable targeted attacks.
\label{sec:experiments}
\subsection{Experimental Setup}
\subsubsection{Dataset}
We use an ImageNet-compatible dataset\footnote{\url{https://github.com/cleverhans-lab/cleverhans/blob/11ea10/examples/nips17_adversarial_competition/dataset/dev_dataset.csv}} containing 1,000 images provided by the NIPS 2017 adversarial attack competition \cite{NIPS2017_competition}.
Each image in the dataset has an officially assigned target class for fair comparison.

\subsubsection{Networks}
We perform experiments on four naturally trained CNNs: Inception-v3 (Inc-v3) \cite{CVPR_InceptionV3}, ResNet-50 (Res-50) \cite{CVPR_ResNet}, VGGNet-16 (VGG-16) \cite{ICLR_VGG} and DenseNet-121 (Den-121) \cite{CVPR_DenseNet}, four naturally trained Vision Transformers (ViTs): ViT-Small-Patch16-224 (ViT-S-16-224), ViT-Base-Patch16-224 (ViT-B-16-224)\cite{vit_ICLR2021}, Swin-Small-Patch4-Window7-224 (Swin-S-224), Swin-Base-Patch4-Window7-224 (Swin-B-224)\cite{Swin_ICCV_2021}, three naturally trained Multi-Layer Perceptrons (MLPs): Mixer-Base-Patch16-224 (MLP-Mixer) \cite{MLPMixer_NIPS2021}, ResMLP-Layer24-224 (ResMLP) \cite{ResMLP}, gMLP-Small-Patch16-224 (gMLP) \cite{gMLP_NIPS2021}, and two adversarially trained CNNs: ens3-adv-Inception-v3 (Inc-v3ens3) and ens-adv-inception-resnet-v2 (IncRes-v2ens) \cite{ICLR_ens_AdvTraining}.
All the networks are publicly accessible.
\subsubsection{Hyper-parameters}
Our method includes three input transformations TI, DI, and SI. Following \cite{CVPR_Poincare}, we set the probability $p_{\textrm{DI}}$ of DI to be $0.7$ and select a Gaussian kernel with a kernel length of $5$ for $\textit{\textbf{W}}$ in TI. For SI, due to the limited computing resources, we set the number of scale copies $M=3$.
Following \cite{BoostingMomentum, ICLR_Nesterov, CVPR_Poincare, NIPS_Simple}, the momentum decay factor $\mu$ is set to $1$.
For all the iterative attacks in the experiments, we use $100$ iterations with learning rate $\alpha=2/255$ as in \cite{NIPS_Simple}.
We use the perturbation budget $\epsilon=16$ under $L_{\infty}$ norm in all the experiments, complying with the rule in the NIPS 2017 competition.
Last but not least, for our proposed DWP, the probability $p_{\textrm{bern}}$ is $0.5$ and the prunable rate $r$ is $0.7$. In other words, we prune $35\%$ of the connections of each network in expectation at each iteration.
\subsubsection{Baseline Methods}
We compare the targeted transferability of DWP and the previous model augmentation methods, Ghost Networks (\ghost{}) and Dual-Stage Network Erosion (\dual{}) \cite{AAAI_Ghost, dual_erosion}, in combination with the state-of-the-art techniques NI-SI-TI-DI. For non-residual networks like VGG-16 and Inc-v3, we insert dropout layers after each activation function. As for residual networks such as Res-50 and Den-121, we apply skip connection erosion on the blocks of each network. \ghost{} \cite{AAAI_Ghost} drops activation outputs with a dropout rate $\Lambda^{\textrm{drop}}_{\textrm{\ghost{}}}$ and multiplies the skip connection by a factor sampled from the uniform distribution $U[1-\Lambda^{\textrm{skip}}_{\textrm{\ghost{}}},1+\Lambda^{\textrm{skip}}_{\textrm{\ghost{}}}]$. Based on \ghost{}, \dual \cite{dual_erosion} not only uses a dropout rate $\Lambda^{\textrm{drop}}_{\textrm{\dual{}}}$, but also scales the values passing dropout by an uniform random factor from $U[1-\Lambda^{\textrm{scale}}_{\textrm{\dual{}}},1+\Lambda^{\textrm{scale}}_{\textrm{\dual{}}}]$. \dual{} also alters the skip connections like \ghost{} with $U[1-\Lambda^{\textrm{skip}}_{\textrm{\dual{}}},1+\Lambda^{\textrm{skip}}_{\textrm{\dual{}}}]$, and introduces an additional bias factor $\gamma$. We set $\Lambda^{\textrm{drop}}_{\textrm{\ghost{}}}=0.012, \Lambda^{\textrm{skip}}_{\textrm{\ghost{}}}=0.22, \Lambda^{\textrm{drop}}_{\textrm{\dual{}}}=0.01, \Lambda^{\textrm{scale}}_{\textrm{\dual{}}}=0.1, \Lambda^{\textrm{skip}}_{\textrm{\dual{}}}=0.14,$ and $\gamma=0.8$ following \cite{AAAI_Ghost, dual_erosion}.

\subsection{Transferable Targeted Attack in Various Scenarios}
We consider targeted transferability under four scenarios: transferring across CNNs, transferring to adversarially trained models, Non-CNN architectures, and the real-world Google Cloud Vision service. We prepare specified networks for each case.
We generate adversarial examples on the ensemble of the white-box models and evaluate targeted success rates on the specified black-box model.
No access to the black-box model is allowed during an attack. Note that for the ensemble, we use equal weights $\beta_k=1/K$ for each of the $K$ white-box models.

\subsubsection{Transferring across Naturally Trained CNNs}
As convolution neural networks are widely used, we first examine the case between CNNs. We select the four classic CNN networks following \cite{NIPS_Simple}: Res-50, VGG-16, Den-121, and Inc-v3.

Table~\ref{tab:Transfer-across-CNNs-Targeted-Success-Rate} shows the results of transferable targeted attacks between CNNs. To inspect the compatibility of DWP, we apply DWP on each of the components in \baseline{}. DWP boosts almost all the attack methods and outperforms \ghost{} and \dual{} in combination with \baseline{}.
As the four CNNs possess designs such as Residual, Dense, and Inception blocks, the results demonstrate the benefits of the diversified ensemble in attacking black-box CNNs with different mechanisms from the white-box substitutes. Protecting necessary connections is also shown to be advantageous.

\begin{table}[t]
\centering
\resizebox{1\columnwidth}{!}{
\begin{tabular}{l| l l l l || l}
\hline
Attack & -Res-50     & -Den-121  & -VGG16        & -Inc-v3       & Avg\\
\hline
TI          & 24.1  & 26.2  & 31.8  & 8.30  & 22.60 \\
+DWP        & 39.9  & 41.7  & 58.3  & 16.8  & \textbf{39.18} \\
\hline
DI          & 66.0  & 75.7  & 76.1  & 48.6  & 66.60\\
+DWP        & 65.2  & 78.0  & 82.9  & 51.7  & \textbf{69.45}\\
\hline
NI          & 15.2  & 17.6  & 17.6  & 6.10&  14.13\\
+DWP        & 35.1  & 33.7  & 50.5  & 12.6  & \textbf{32.98}\\
\hline
SI          & 37.6  & 44.7  & 38.9  & 17.7  & 34.73\\
+DWP        & 51.6  & 63.7  & 68.2  & 28.8  & \textbf{53.23}\\
\hline
\hline
NI-SI-TI-DI & 76.1&  86.7&  77.1&  66.9& 76.70\\
+\ghost{}   & 68.7&  85.0&  80.1&  72.4& 76.55\\
+\dual{}    & 67.0&  75.1&  79.1&  66.7& 71.98\\
+DWP        & 77.7&  89.4&  87.2&  70.9& \textbf{81.30}\\
\hline
\end{tabular}
}
\caption{The targeted success rates of transferring across CNNs. The ``-'' prefix stands for the black-box network with the other three serving as the white-box ones for ensemble. ``+'' means the participation of a specific model augmentation method. DWP outperforms other leading model augmentation methods \ghost{} and \dual{}.}
\label{tab:Transfer-across-CNNs-Targeted-Success-Rate}
\end{table}

\subsubsection{Transferring to Adversarially Trained Models}
Adversarial training \cite{ICLR_ens_AdvTraining, ICLR_MadryPGDAdvTraining} is one of the primary techniques for defending against malicious attacks. It brings robustness to models by training them with adversarial examples. Under the scenario of transferring to adversarially trained models, we ensemble only the four naturally trained networks (Res-50, Den-121, VGG16, and Inc-v3) as white-box models to simulate the situation where attackers have few details about defense. The two adversarially trained networks (Inc-v3ens3 and IncRes-v2ens) will act as our black-box model separately.

Table~\ref{tab:Transfer-to-Robust-Models-Targeted-Success-Rate} summarizes the results of transferring to adversarially trained networks. Targeted success rates under this scenario are lower due to the robustness of adversarially trained models. Under such a challenging scenario, DWP still helps alleviate the discrepancy between white-box naturally trained and black-box adversarially trained networks, bringing about up to $10.1$\% improvement on average. The power of the diversified ensemble under the premise of protecting necessary connections is highlighted again, especially for black-box networks with significant differences from the white-box ones.

\subsubsection{Transferring to Non-CNN Architectures}
In practice, information about the networks used by defenders remains unknown to attackers. A targeted attack method is more practical if adversarial examples can transfer to black-box architectures different from the white-box ones accessible by attackers. Beyond CNNs, recent works attempt to solve computer vision tasks using Vision Transformers (ViTs) \cite{vit_ICLR2021, Swin_ICCV_2021} and Multi-Layer Perceptrons (MLPs) \cite{MLPMixer_NIPS2021, gMLP_NIPS2021, ResMLP}. To be more comprehensive, we evaluate the targeted transferability from CNNs to these architectures.

We generate targeted adversarial images on the ensemble of the four naturally trained CNNs. \baseline{} comes with the three model augmentation methods, respectively, including our DWP. From Table~\ref{tab:Transfer-to-Non-CNN-Architectures-Targeted-Success-Rate}, model augmentations are effective even though the black-box networks have no convolution operations other than the input projection. Our DWP improves the results on both ViTs and MLPs, outperforming all the other methods.

\begin{table}[t]
    \centering
    \begin{tabular}{l| c c c c}
    \hline
    Attack Method & NI-SI-TI-DI & +\ghost{} & +\dual{} & +DWP \\
    \hline
    Inc-v3ens3    & 50.0& 51.6& 49.7& \textbf{65.3}\\
    IncRes-v2ens  & 19.4& 29.8& 34.5& \textbf{39.0}\\
    \hline
    Average       & 34.7& 40.7& 42.1& \textbf{52.15}\\
    \hline
    \end{tabular}
    \caption{The targeted success rates of transferring to adversarially trained networks. Our DWP outperforms \ghost{} and \dual{} over 10\%.}
    \label{tab:Transfer-to-Robust-Models-Targeted-Success-Rate}
\end{table}

\begin{table}[t]
    \centering
    \begin{tabular}{l| c c c c}
    \hline
    Attack Method & NI-SI-TI-DI & +\ghost{} & +\dual{} & +DWP \\
    \hline
    ViT-S-16-224     & 25.9& 31.5& 31.1& \textbf{37.3}\\
    ViT-B-16-224     & 24.8& 29.9& 28.4& \textbf{37.4}\\
    Swin-S-224       & 26.7& 29.1& 26.5& \textbf{36.7}\\
    Swin-B-224       & 23.9& 27.1& 23.9& \textbf{32.9}\\
    MLP-Mixer        & 21.7& 24.2& 27.0& \textbf{30.9}\\
    ResMLP           & 51.3& 56.5& 54.4& \textbf{64.1}\\
    gMLP             & 20.4& 25.3& 26.9& \textbf{30.4}\\
    \hline
    Average          & 27.81 & 31.94 & 31.17 & \textbf{38.53}\\
    \hline
    \end{tabular}
    \caption{The targeted success rates of transferring to Non-CNN architectures. Our DWP maintains higher success rates stably.}
    \label{tab:Transfer-to-Non-CNN-Architectures-Targeted-Success-Rate}
\end{table}

\subsubsection{Transferring to Google Cloud Vision}
For a more practical scenario, we use Google Cloud Vision to evaluate our targeted adversarial examples. Google Cloud Vision predicts a list of labels with their corresponding confidence scores and only returns label annotations with confidence above 50\%. The scenario is completely black-box since no information about gradients and parameters of the underlying system is accessible. Previous works leverage query-based attacks \cite{Decision_ICLR2018, Query_ICML2018, Devil_USENIX2020} or black-box transferability \cite{DelvingLiu2017, NIPS_Simple}. However, query-based methods often require large numbers of queries, and the existing transferable attacks still have substantial room for improvement.

In this experiment, we randomly select $100$ images correctly labeled by Google Cloud Vision from the Imagenet-compatible dataset. Similarly, we use the four naturally trained CNNs to generate adversarial examples. We identified an image as a successful attack if at least one of the labels returned by Google Cloud Vision is semantically close to its corresponding target class. We summarize the attack results in Table~\ref{tab:GVision}. DWP outperforms the previous model augmentation methods by 7\%.

\begin{table}
\centering
\begin{tabular}{|l| l | l | l |}
\hline
NI-SI-TI-DI & +\ghost{} & +\dual{} & +DWP\\
\hline
27 & 43 & 42 & \textbf{50}\\
\hline
\end{tabular}
\caption{Targeted success rates (\%) on Google Cloud Vision}
\label{tab:GVision}
\end{table}

\begin{figure}
\centering
\begin{subfigure}{0.45\columnwidth}
    \centering
    \includegraphics[width=\textwidth]{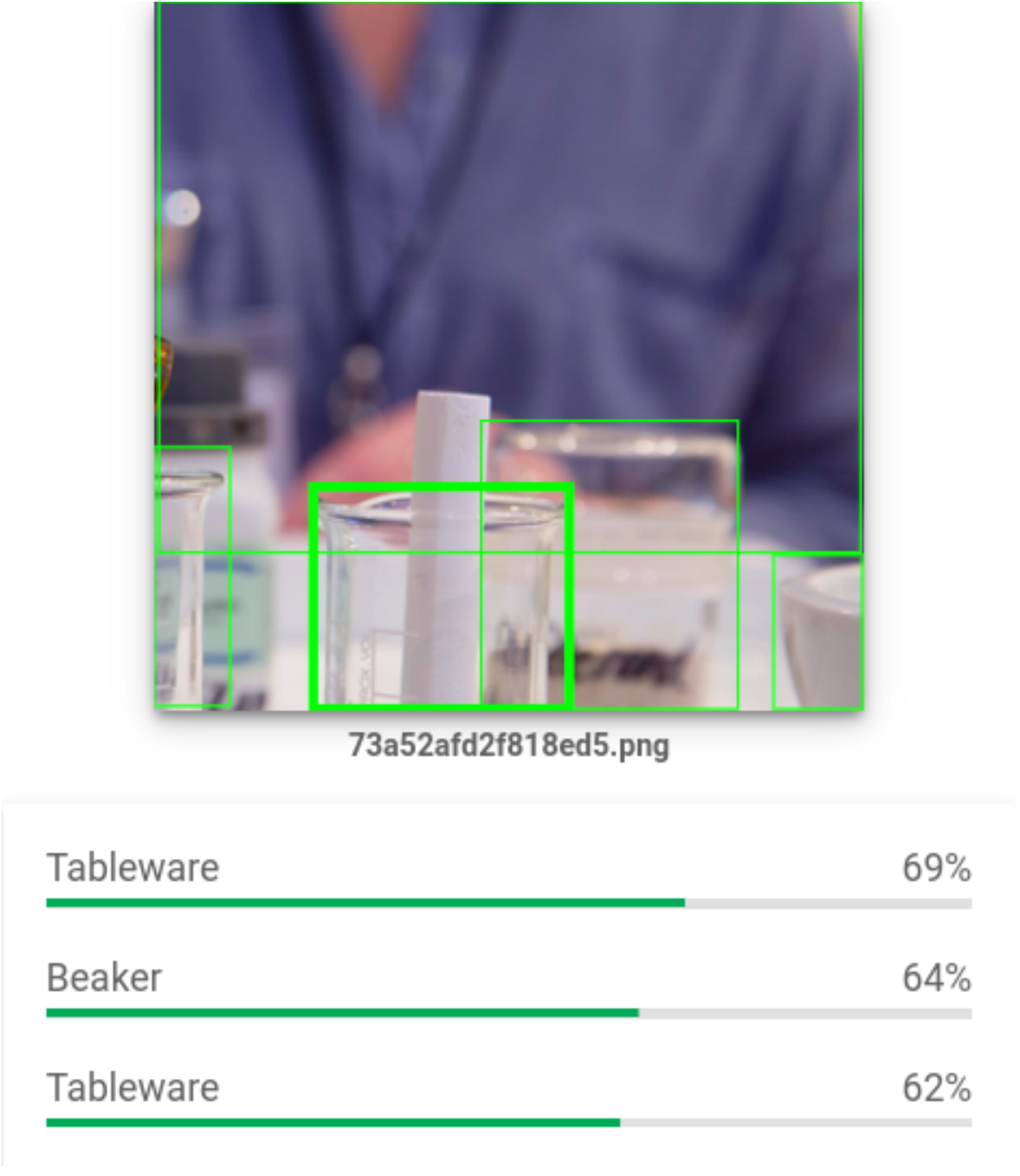}
    \caption{Clean image}
    \label{fig:gcv1}
\end{subfigure}
\begin{subfigure}{0.45\columnwidth}
    \centering
    \includegraphics[width=\textwidth]{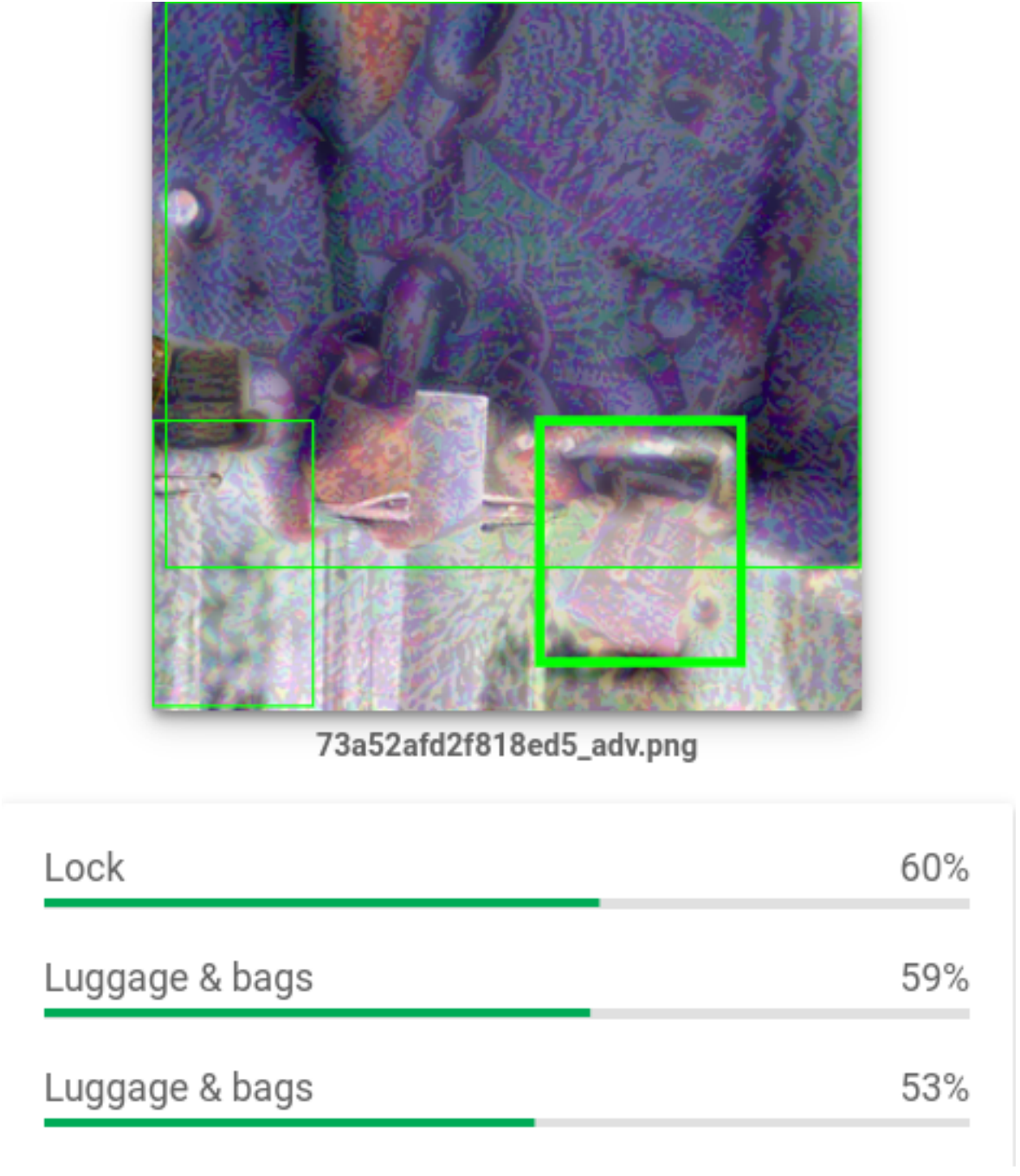}
    \caption{NI-SI-TI-DI + DWP}
    \label{fig:gcv2}
\end{subfigure}
\caption{A demo of our DWP attack on Google Cloud Vision. The attacked image with the ground truth label of ``Beakers'' is recognized as the target class ``Padlocks'' assigned by the NIPS 2017 Imagenet-compatible dataset.}
\vspace{-2mm}
\end{figure}

\subsection{Perturbations from Different Pruned Models}
To investigate whether our method can promote diversity of the ensemble, we examine the relationship between the generated adversarial perturbations instead of between model outputs. The reason is that compared to the logit outputs by models, perturbation vectors are more direct in determining the update of adversarial examples.
Recent works \cite{ICLR_Nesterov, CVPRTI} have proposed methods improving transferability with output-preserving operations. Although these operations retain model outputs, they modify the gradients and enrich the directions of adversarial perturbations.
With these motivations, we focus on the diversity between perturbations computed from pruned models.

Liu et al. \cite{DelvingLiu2017} first studied the effectiveness of ensemble in enhancing transferability. 
They demonstrate the diversity of the ensemble by showing near-zero cosine similarities between perturbations from different white-box networks.
Following \cite{DelvingLiu2017}, we calculate cosine similarities between perturbations generated from the additional \newborn{} models produced by DWP. From each of our four naturally trained CNNs, we acquire five \newborn{} models with different connections pruned. We term the cosine similarity between perturbations of pruned models from an identical CNN as intra-CNN similarity. The case from different CNNs is termed as inter-CNN similarity. To avoid cherry-picking, both intra-CNN and inter-CNN similarities come from the average of the first ten images in the ImageNet-compatible dataset. Futhermore, we only use NI in combination with DWP to produce perturbations in this experiment to prevent the influence of factors other than pruning.

\begin{figure}
\centering
\includegraphics[width=0.9\columnwidth]{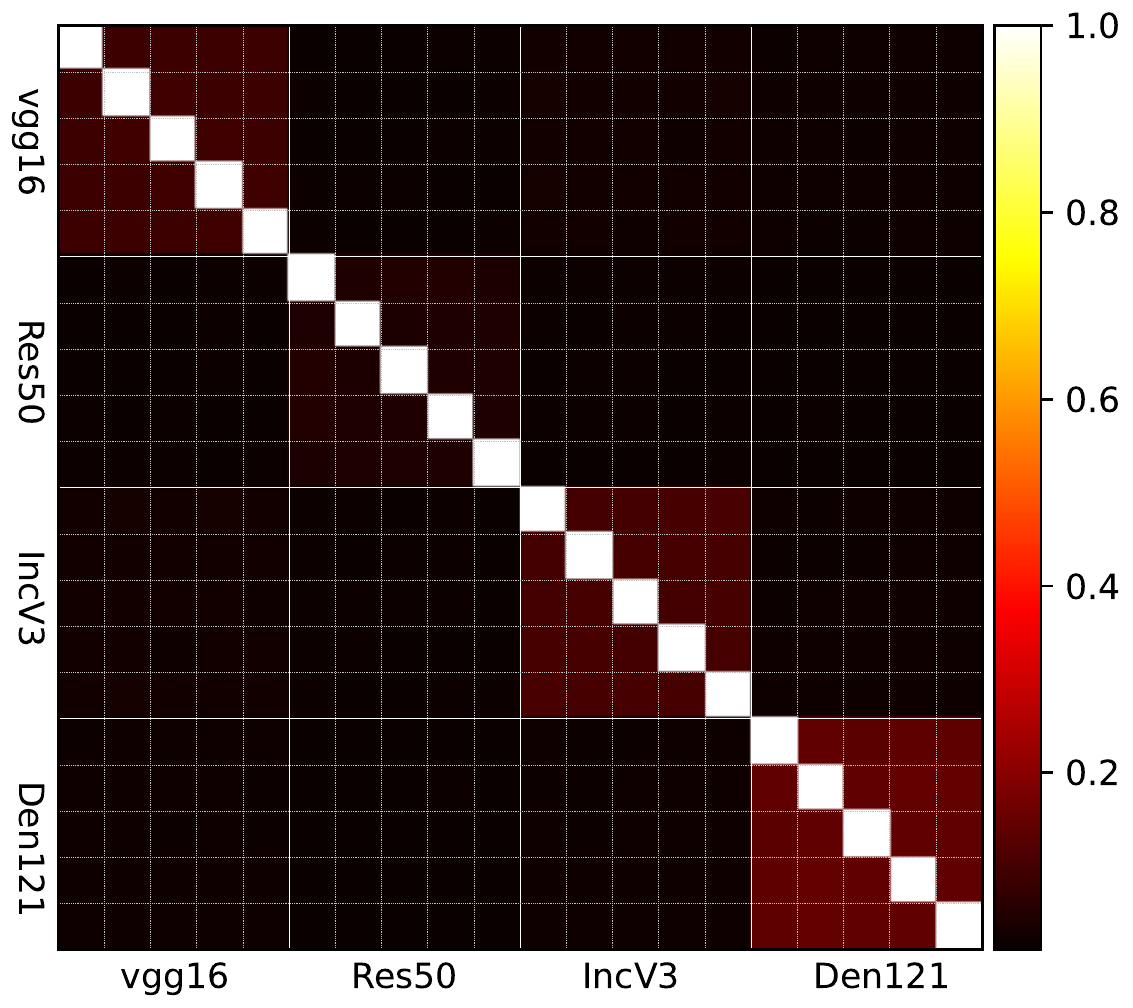}
\caption{Perturbation cosine similarities between \newborn{} models. Each diagonal block summarizes 10 ($C^5_2$) intra-CNN similarity cells. Each non-diagonal block summarizes 25 ($5\times5$) inter-CNN similarity cells. The pairwise cosine similarity matrix is symmetric and shows orthogonality between perturbations.}
\label{fig:ortho_newborn}
\vspace{-2mm}
\end{figure}

Figure~\ref{fig:ortho_newborn} is a symmetric matrix containing $16$ ($4\times4$) blocks. The diagonal blocks summarize ten ($C^5_2$) intra-CNN similarities while the non-diagonal blocks summarize 25 ($5\times5$) inter-CNN similarities in cells. The diagonal cells are all $1.0$ since they are all from two identical perturbation vectors. As for the non-diagonal cells, we find the cell values in diagonal blocks (intra-CNN) slightly higher than in non-diagonal blocks (inter-CNN). However, these values are still close to zero, appearing dark red. The results show that whether two \newborn{} models come from the same CNN, the generated perturbations generated are always nearly orthogonal.
These observations on orthogonality support our claim that \newborn{} models obtained via DWP provide more diversity for attacking.

\subsection{Semantics of the Target Class}
\begin{figure*}
\centering
\captionsetup[subfigure]{labelformat=empty}
\begin{subfigure}{\demoImgWidth{}}
    \centering
    \includegraphics[width=\textwidth]{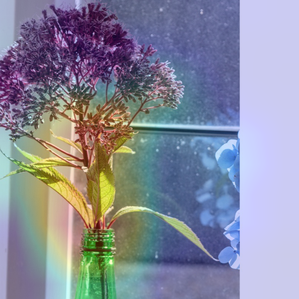}
    \caption{Clean image \textbf{Vase}}
    \label{fig:img11}
\end{subfigure}
\begin{subfigure}{\demoImgWidth}
    \centering
    \includegraphics[width=\textwidth]{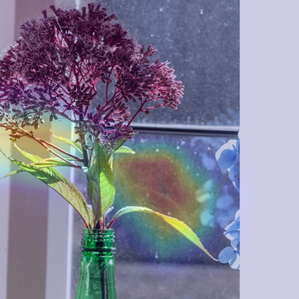}
    \caption{Clean image \textbf{Marmoset}}
    \label{fig:img12}
\end{subfigure}
\begin{subfigure}{\demoImgWidth}
    \centering
    \includegraphics[width=\textwidth]{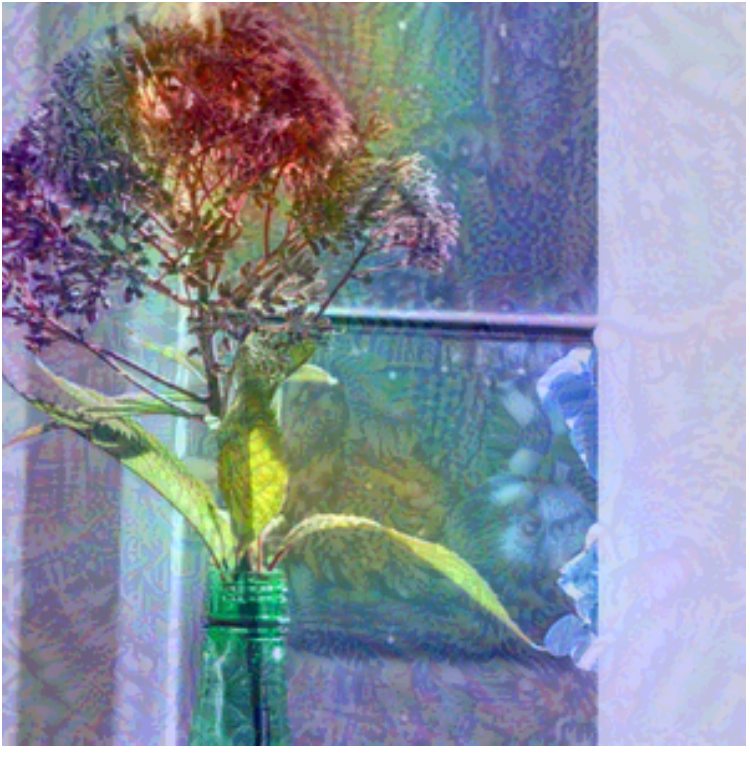}
    \caption{NI-SI-TI-DI \textbf{Marmoset}}
    \label{fig:img13}
\end{subfigure}
\begin{subfigure}{\demoImgWidth}
    \centering
    \includegraphics[width=\textwidth]{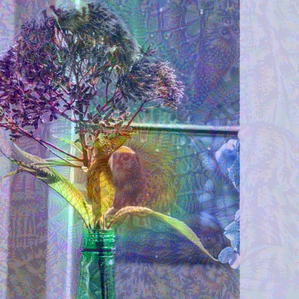}
    \caption{+\ghost{} \textbf{Marmoset}}
    \label{fig:img14}
\end{subfigure}
\begin{subfigure}{\demoImgWidth}
    \centering
    \includegraphics[width=\textwidth]{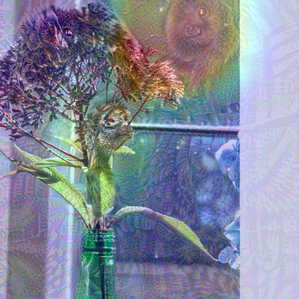}
    \caption{+\dual{} \textbf{Marmoset}}
    \label{fig:img15}
\end{subfigure}
\begin{subfigure}{\demoImgWidth}
    \centering
    \includegraphics[width=\textwidth]{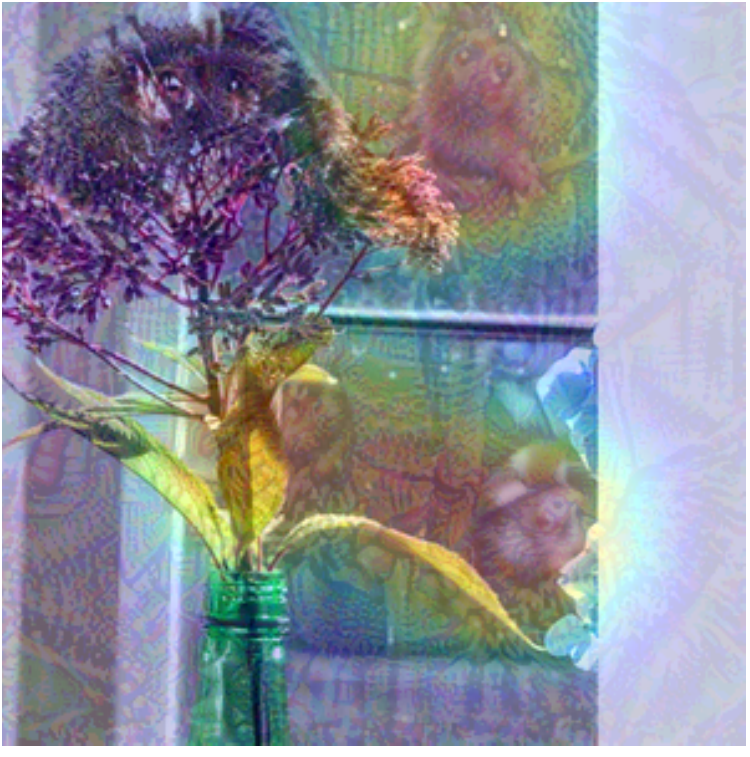}
    \caption{+DWP \textbf{Marmoset}}
    \label{fig:img16}
\end{subfigure}

\begin{subfigure}{\demoImgWidth{}}
    \centering
    \includegraphics[width=\textwidth]{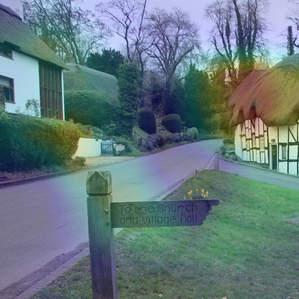}
    \caption{Clean image \textbf{Thatch}}
    \label{fig:img21}
\end{subfigure}
\begin{subfigure}{\demoImgWidth}
    \centering
    \includegraphics[width=\textwidth]{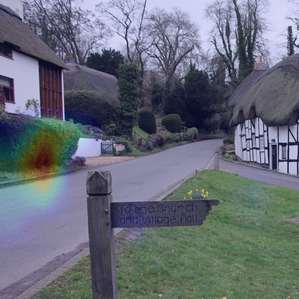}
    \caption{Clean image \textbf{Poodle}}
    \label{fig:img22}
\end{subfigure}
\begin{subfigure}{\demoImgWidth}
    \centering
    \includegraphics[width=\textwidth]{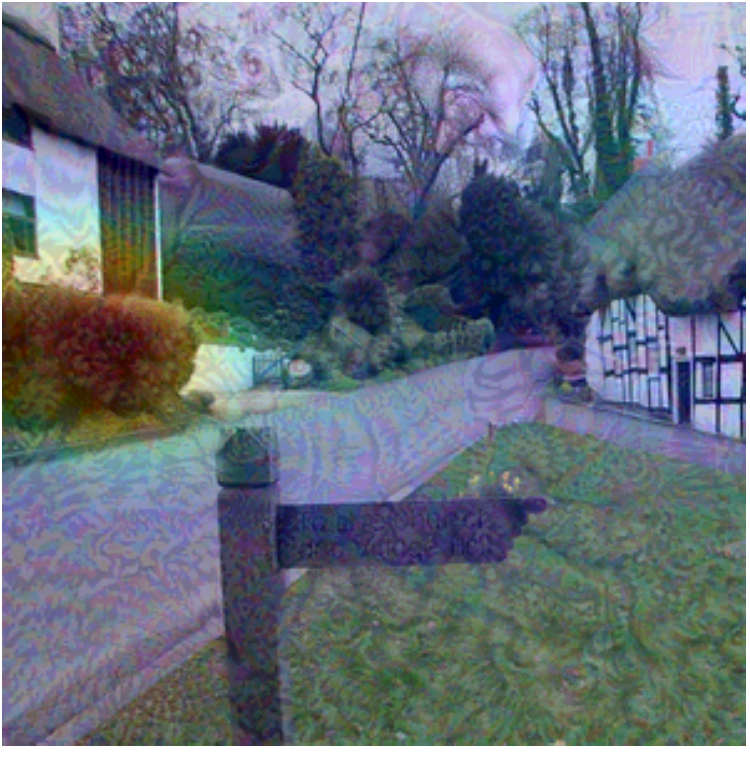}
    \caption{NI-SI-TI-DI \textbf{Poodle}}
    \label{fig:img23}
\end{subfigure}
\begin{subfigure}{\demoImgWidth}
    \centering
    \includegraphics[width=\textwidth]{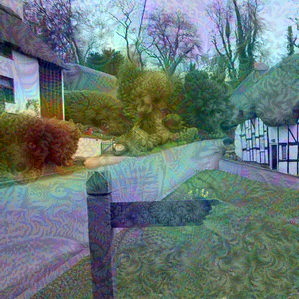}
    \caption{+\ghost{}  \textbf{Poodle}}
    \label{fig:img24}
\end{subfigure}
\begin{subfigure}{\demoImgWidth}
    \centering
    \includegraphics[width=\textwidth]{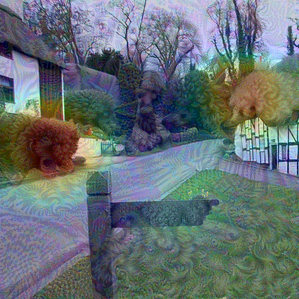}
    \caption{+\dual{}  \textbf{Poodle}}
    \label{fig:img25}
\end{subfigure}
\begin{subfigure}{\demoImgWidth}
    \centering
    \includegraphics[width=\textwidth]{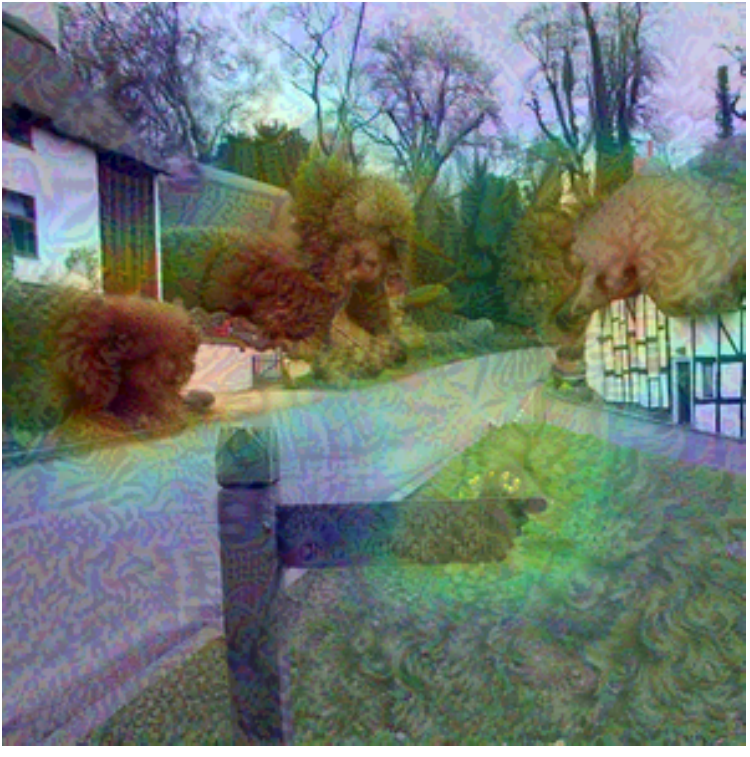}
    \caption{+DWP  \textbf{Poodle}}
    \label{fig:img26}
\end{subfigure}
\caption{The GradCAMs of clean and targeted adversarial images on naturally trained ResNet-50. The two leftmost columns show GradCAMs of clean images regarding the ground truth and target class. The other columns provide the GradCAMs of adversarial images generated by different methods. Targeted perturbations guide the highlighted area and impose semantics of the target class on images.}
\label{fig:target-semantics}
\end{figure*}

Prior work has shown that targeted adversarial examples semantically close to the target class tend to be more transferable \cite{ICCV2021_GAN, FDA_ICLR2020, FDAxent_NIPS2020}. To provide more insight into the success of DWP, we explore the patterns in targeted adversarial examples.

GradCAM \cite{GradCAM_ICCV2017} uses the mean gradient values of a specific class output confidence with respect to each intermediate feature map to be its corresponding coefficient. The weighted average of feature maps using these coefficients provides an explanation of a particular decision made by the model. In Figure~\ref{fig:target-semantics}, we draw GradCAMs on naturally trained ResNet-50 to provide some explainable observations on adversarial images generated by different methods.

In the two leftmost columns, we show the GradCAMs of clean images with their ground truth and target class, respectively. GradCAMs correctly highlight regions about the ground truth class on the clean images. On the other hand, without adversarial perturbations, there is no evident relation between the target class and the corresponding highlighted areas.
The other four columns show the GradCAMs with the target class of adversarial images generated by \baseline{} and \baseline{} plus \ghost{}, \dual{}, and DWP, respectively. The adversarial perturbations produce target class-specific patterns and guide the highlighted region of GradCAMs. Note that the perturbation budget is limited to $l_{\infty}\leq16$ to ensure the attacks are quasi-imperceptible.

\begin{figure}
\centering
\includegraphics[width=\columnwidth]{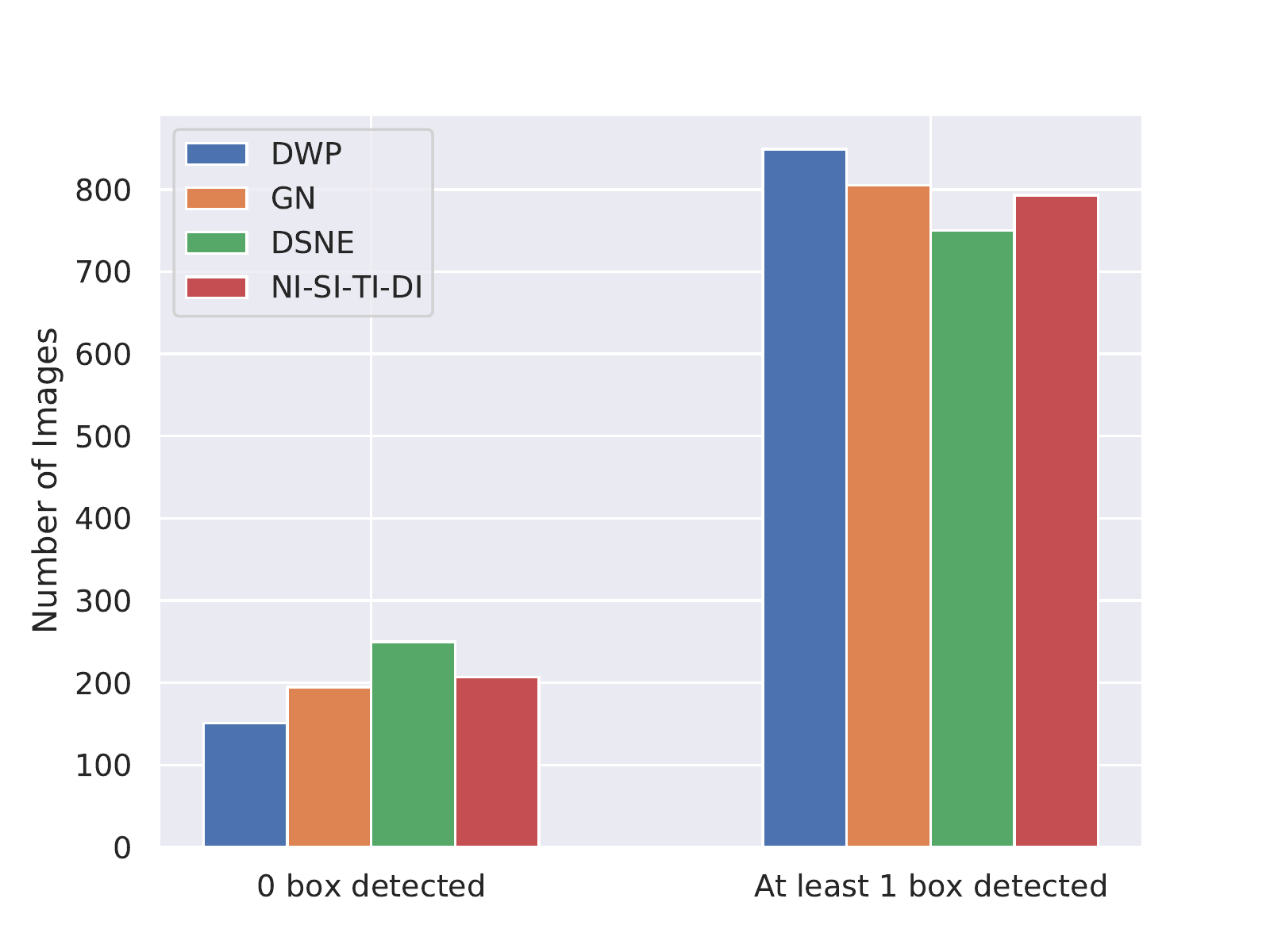}
\caption{The comparison of object detection results of the target class. Compared to \ghost{} and \dual{}, adversarial images generated by our DWP are more likely to contain at least one object detected as the target class.}
\label{fig:bbox-counts}
\vspace{-5mm}
\end{figure}

For quantitative comparison, we leverage an object detector\footnote{\url{https://github.com/ibaiGorordo/ONNX-ImageNet-1K-Object-Detector}} to detect target-class patterns in the targeted adversarial images. We set the threshold to 0.1, which is lower than usual, to capture more potential patterns in images. The bars on the right side of Figure~\ref{fig:bbox-counts} summarize the number of images with at least one bounding box detected. Compared to other methods, DWP is the method most likely to generate adversarial examples with at least one target-class pattern detected. Notice that under the limit of the perturbation budget, even though we have a lower threshold, there are still about 150 images without any target-class object detected according to the bars on the left side of Figure~\ref{fig:bbox-counts}. Since we do not integrate object detectors into our attack procedure, the results support that our DWP is better at producing target-class-specific information.

\section{Conclusion}
In this paper, we propose Diversified Weight Pruning (DWP) leveraging network compression to improve the targeted transferability of adversarial attacks. DWP produces additional pruned models for ensemble via weight pruning. Due to the over-parameterized property, the quality of pruned models introduced by DWP is well-preserved. Experiments show that by protecting the necessary weight connections of networks, targeted adversarial examples are more likely to acquire semantics of the target class. By evaluating DWP on ImageNet, we show that DWP improves the state-of-the-art model augmentation methods on transferable targeted attacks, especially for challenging scenarios such as transferring to adversarially trained models and Non-CNN architectures. We hope that our work can serve as a bridge between network compression and transferable attack, inspiring more collaboration.

{\small
\bibliographystyle{ieee_fullname}
\bibliography{egbib}
}

\newpage
\clearpage
\appendix
\section*{Appendix}
\section{Ablation Analysis on Prunable Rates}
In the ablation analysis, we explore targeted attack success rates under different prunable rates $r$. As the prunable rate determines the number of connections possible to be pruned during attacking, white-box models can produce more diverse \newborn{} models using higher prunable rates. However, with excessive connections pruned away, the quality of \newborn{} networks will be unstable.

\begin{figure}[ht]
\centering
\includegraphics[width=\columnwidth]{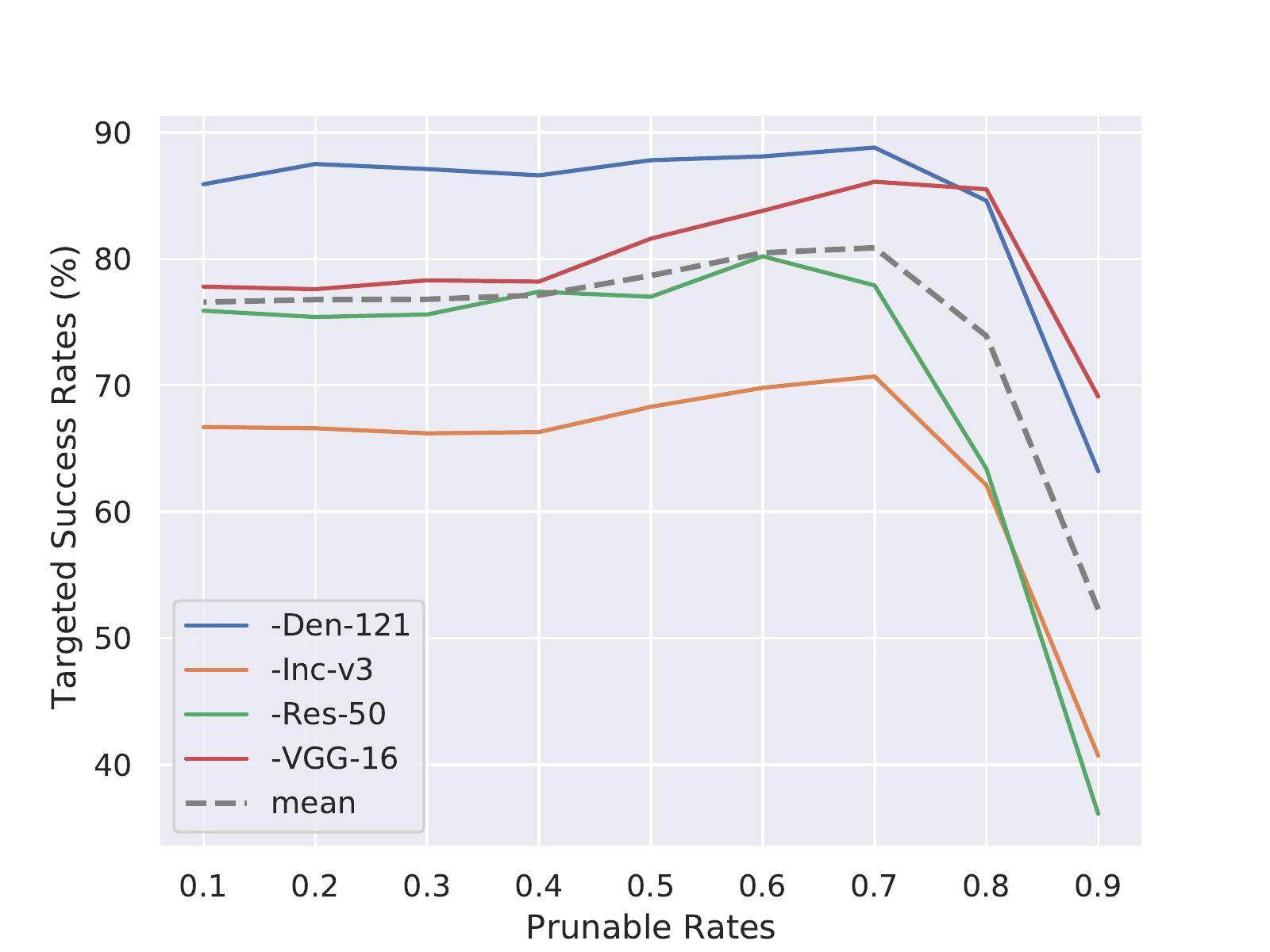}
\caption{\textbf{The targeted success rates under different prunable rates $r$ on each black-box model.} Each curve shows the trade-off between the diversity and stability of \newborn{} models. The curve for mean targeted success rates reaches its maximum at $r=0.7$.}
\label{fig:ablation-prunable-rates}
\end{figure}

\begin{figure}[ht]
\centering
\includegraphics[width=\columnwidth]{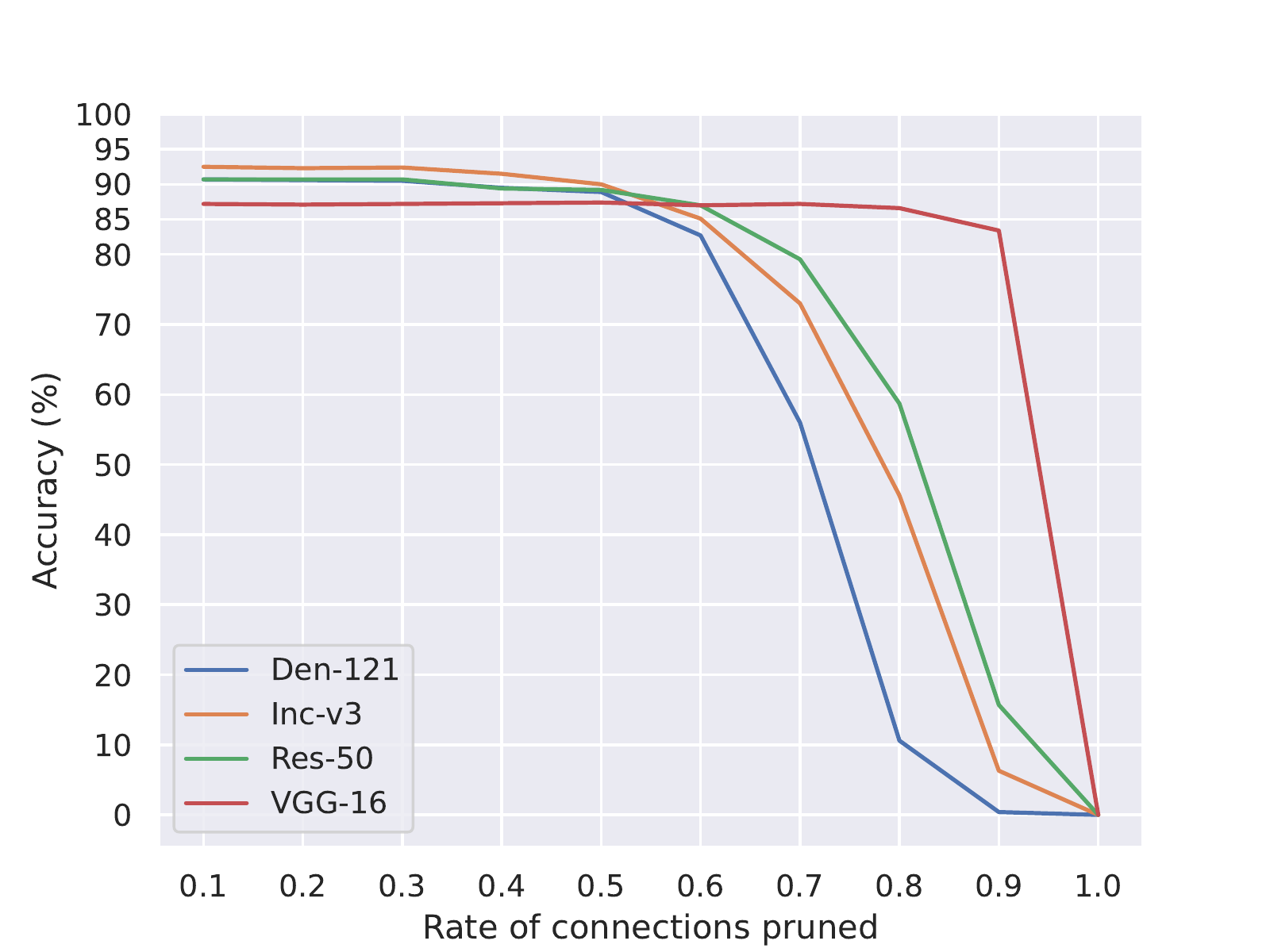}
\caption{\textbf{The decay on the accuracy of each network with respect to the rate of minor weight connections pruned.}}
\label{fig:acc-prune}

\end{figure}

To find the sweet spot to the trade-off, we enumerate different prunable rates, conducting the attack experiments with all the other hyper-parameters as default. 
Figure~\ref{fig:ablation-prunable-rates} shows the trade-off.  
We select $r=0.7$ throughout our experiments as the curve of mean targeted success rates reaches its maximum.
With our designated prunable rate $r=0.7$, DWP prunes about $35\%$ of weight connections. Figure~\ref{fig:acc-prune} shows the decline in the accuracy of the four CNNs with different rates of minor weight connections pruned.

\section{Transferring across CNNs with Similar Architectures}
Table~\ref{tab:prev-CNNs-Targeted-Success-Rate} summarizes the targeted attack success rates across Inception-v3 (Inc-v3), Inception-v4 (Inc-v4), Inception-Resnet-v2 (IncRes-v2) \cite{AAAI_IncV4_IncResV2} and ResNet-101 (Res-101) \cite{CVPR_ResNet}. The group of CNNs was popular for evaluating attacks \cite{CVPR_red_var_ensemble, CVPRTI, CVPRDI, CVPR_Poincare, BoostingMomentum}. However, due to similar architectures between these CNNs, \cite{NIPS_Simple} suggests using a group of networks with relatively diverse designs. For completeness, we also include the targeted success rates of different model augmentation methods under this group of similar CNNs.

\begin{table}[h]
\centering
\resizebox{\columnwidth}{!}{
\begin{tabular}{l| l l l l || l}
\hline
Attacks & -Inc-v3 & -Inc-v4  & -IncRes-v2 & -Res-101 & Avg\\
\hline
NI-SI-TI-DI  &  65.2 &  71.3 & 73.2 &  20.9 & 57.65 \\
+\ghost{}    &  77.5 &  70.0 & 69.0 &  26.1 & 60.65\\
+\dual{}     &  70.7 &  60.3 & 69.5 &  13.7 & 53.55\\
+DWP         &  \textbf{83.1} &  \textbf{86.1} & \textbf{85.4} & \textbf{40.6} & \textbf{73.80}\\
\hline
\end{tabular}
}
\caption{The targeted success rates of transferring across similar CNN architectures. The ``-'' prefix stands for the black-box network with the other three serving as the white-box ones for the ensemble. ``+'' means the participation of a specific model augmentation method. DWP outperforms other leading model augmentation methods \ghost{} and \dual{}.}
\label{tab:prev-CNNs-Targeted-Success-Rate}
\end{table}

\section{Transferring to Multi-Step Adversarially Trained Models}
Authors in \cite{ICLR_ens_AdvTraining} propose ``ensemble adversarial training'', which trains the network with adversarial examples generated from external models. While the single-step attack in the procedure is less costly, the models fall short of resisting iterative attacks even in black-box scenarios. Thus, we explore the black-box targeted attack results on the models with multi-step adversarial training.

Transferable targeted attacks from naturally trained CNNs to multi-step adversarially trained networks remain an open problem. Recent attacks only show the non-targeted results \cite{NIPS2022_ReversePerturbation}. Even the resource-intensive attack \cite{ICCV2021_GAN} fails to achieve satisfied targeted success rates. We borrow the adversarially trained networks from \cite{DoTransferBetter_NIPS2020} in the following experiments. Table~\ref{tab:Transfer-to-Robust-Models-From-Nature-Targeted-Success-Rate} shows the failure of transferring targeted attacks from the four naturally trained CNNs to various three-step adversarially trained models.

Despite the frustrating results, there is a different story when we generate adversarial examples on multi-step adversarially trained networks. Even if the victim network undergoing multi-step adversarial training has a different architecture, it remains vulnerable to these attacks. Table~\ref{tab:Transfer-to-MultiStep-Models-Targeted-Success-Rate} summarizes the targeted attack results of the ensemble composed of Res-18 ($|\epsilon|_\infty=2$), Res-50 ($|\epsilon|_\infty=2$) and WideRes-50-2 ($|\epsilon|_\infty=2$). The two upper groups in Table~\ref{tab:Transfer-to-MultiStep-Models-Targeted-Success-Rate} report the targeted success rates on different CNN architectures and the norm of $\epsilon$ used in adversarial training. We also provide the results on naturally trained CNNs and the ones with ensemble adversarial training. Our DWP stably benefits the results.

\begin{table}[h]
    \centering
    \resizebox{\linewidth}{!}{
    \begin{tabular}{l| c c c c}
    \hline
    Attack Method & NI-SI-TI-DI & +\ghost{} & +\dual{} & +DWP \\
    \hline
    Res-18 ($|\epsilon|_\infty=1$)          & 0.2 & 0.2 & 0.5 & 0.2\\
    Res-50 ($|\epsilon|_\infty=1$)          & 0.0 & 0.6 & 0.8 & 0.3\\
    WideRes-50-2 ($|\epsilon|_\infty=1$)    & 0.0 & 0.2 & 0.4 & 0.1\\
    \hline
    Res-18 ($|\epsilon|_2=3$)       & 0.0   & 0.1   & 0.1   & 0.0 \\
    Den-121 ($|\epsilon|_2=3$)      & 0.0   & 0.0   & 0.0   & 0.0 \\
    VGG16 ($|\epsilon|_2=3$)        & 0.0   & 0.0   & 0.0   & 0.0 \\
    Resnext-50 ($|\epsilon|_2=3$)   & 0.0   & 0.0   & 0.1   & 0.0 \\
    \hline
    \hline
    \end{tabular}
    }
    \caption{The targeted success rates of transferring to three-step adversarially trained networks from naturally trained CNNs.}
    
    \label{tab:Transfer-to-Robust-Models-From-Nature-Targeted-Success-Rate}
\end{table}

\begin{table}[h]
    \centering
    \resizebox{\linewidth}{!}{
    \begin{tabular}{l| c c c c}
    \hline
    Attack Method & NI-SI-TI-DI & +\ghost{} & +\dual{} & +DWP \\
    \hline
    Res-18 ($|\epsilon|_\infty=1$)          & 33.2 & 33.6 & 21.7 & \textbf{37.0}\\
    Res-50 ($|\epsilon|_\infty=1$)          & 40.5 & 39.4 & 21.0 & \textbf{41.4}\\
    WideRes-50-2 ($|\epsilon|_\infty=1$)    & 37.8 & 35.4 & 18.2 & \textbf{39.5}\\
    \hline
    Res-18 ($|\epsilon|_2=3$)        & 12.6 & 12.6 & 12.6 & \textbf{15.2}\\
    Den-121 ($|\epsilon|_2=3$)       & 17.4 & 18.0 & 11.3 & \textbf{19.2}\\
    VGG16 ($|\epsilon|_2=3$)         & 12.5 & 13.3 & 9.60 & \textbf{15.5}\\
    Resnext-50 ($|\epsilon|_2=3$)    & 19.1 & 19.2 & 11.5 & \textbf{21.0}\\
    \hline
    Res-50     & 21.9 & 16.8 & 12.6 & \textbf{22.3} \\
    Den-121    & 27.6 & 29.0 & 15.9 & \textbf{39.0} \\
    VGG16      & 8.60 & 8.80 & 6.20 & \textbf{18.6} \\
    Inc-v3     & 17.4 & 17.9 & 8.70 & \textbf{26.7} \\
    \hline
    Inc-v3ens3     & 22.4 & 23.3 & 9.30 & \textbf{30.5}\\
    IncRes-v2ens   & 22.3 & 22.6 & 22.4 & \textbf{30.0}\\
    \hline
    \hline
    \end{tabular}
    }
    
    \caption{The targeted success rates of transferring to three-step
adversarially trained networks from the ones with different architectures and $\epsilon$.}
    \label{tab:Transfer-to-MultiStep-Models-Targeted-Success-Rate}
    
\end{table}

\section{In Comparison with SVRE \cite{CVPR_red_var_ensemble}}
\cite{CVPR_red_var_ensemble} proposes Stochastic Variance Reduced Ensemble (SVRE) to improve the naive logit-averaging ensemble. Since SVRE assumes the white-box substitute models remain unchanged throughout attacking, it may not be trivially compatible with the model augmentation methods altering networks at each iteration. Since both techniques aim to enhance the ensemble method, we compare the targeted success rates of SVRE with the model augmentation methods.

\begin{figure}[h!]
    \centering
    \begin{subfigure}{\columnwidth}
    \centering
    \includegraphics[width=\textwidth, trim={0 0mm 0 0cm}, clip]{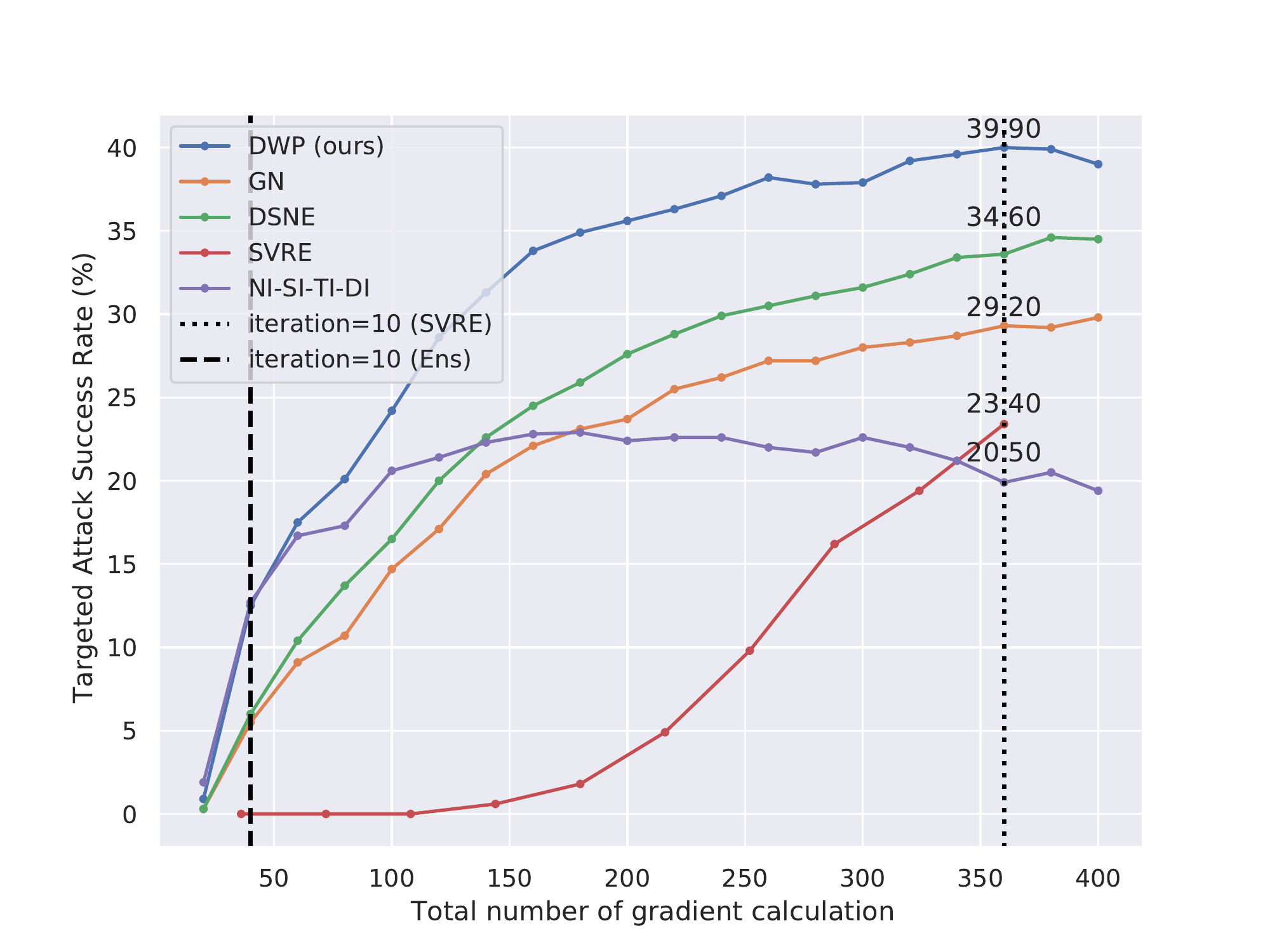}
    \caption{IncRes-v2ens}
    \label{fig:SVRE-IncResEns}
    \end{subfigure}
    
    \begin{subfigure}{\columnwidth}
    \centering
    \includegraphics[width=\textwidth, trim={0 0mm 0 0cm}, clip]{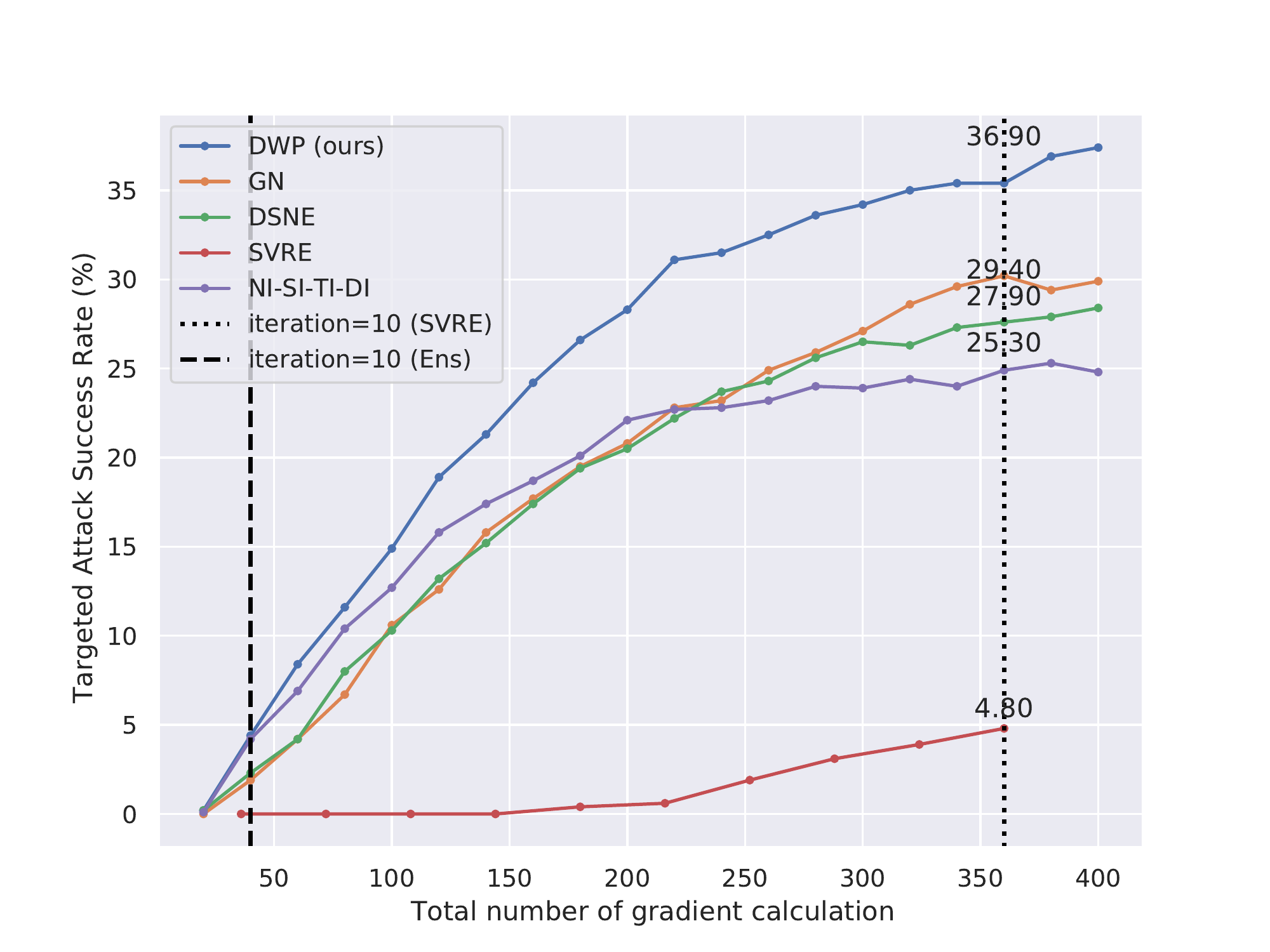}
    \caption{ViT-B-16-224}
    \label{fig:SVRE-VitBase}
    \end{subfigure}

\caption{\textbf{Comparison of targeted transferability between model augmentation methods and SVRE.}}
\label{fig:SVRE}
\end{figure}

We generate the adversarial examples using the four naturally trained CNNs. Figure~\ref{fig:SVRE-IncResEns} and Figure~\ref{fig:SVRE-VitBase} show the comparison between model augmentation methods and SVRE on IncRes-v2ens and ViT-B-16-224, respectively. Since SVRE does nine times more gradient calculations than the naive ensemble at each iteration, we compare the targeted success rates at different numbers of gradient calculations following \cite{CVPR_red_var_ensemble}. SVRE falls short of improving targeted transferability with no additional diversified models introduced.

\newpage
\section{Results of DWP on Google Cloud Vision}
\hspace{-10mm}
\begin{tabular}{ccccc}
  \includegraphics[width=0.18\textwidth]{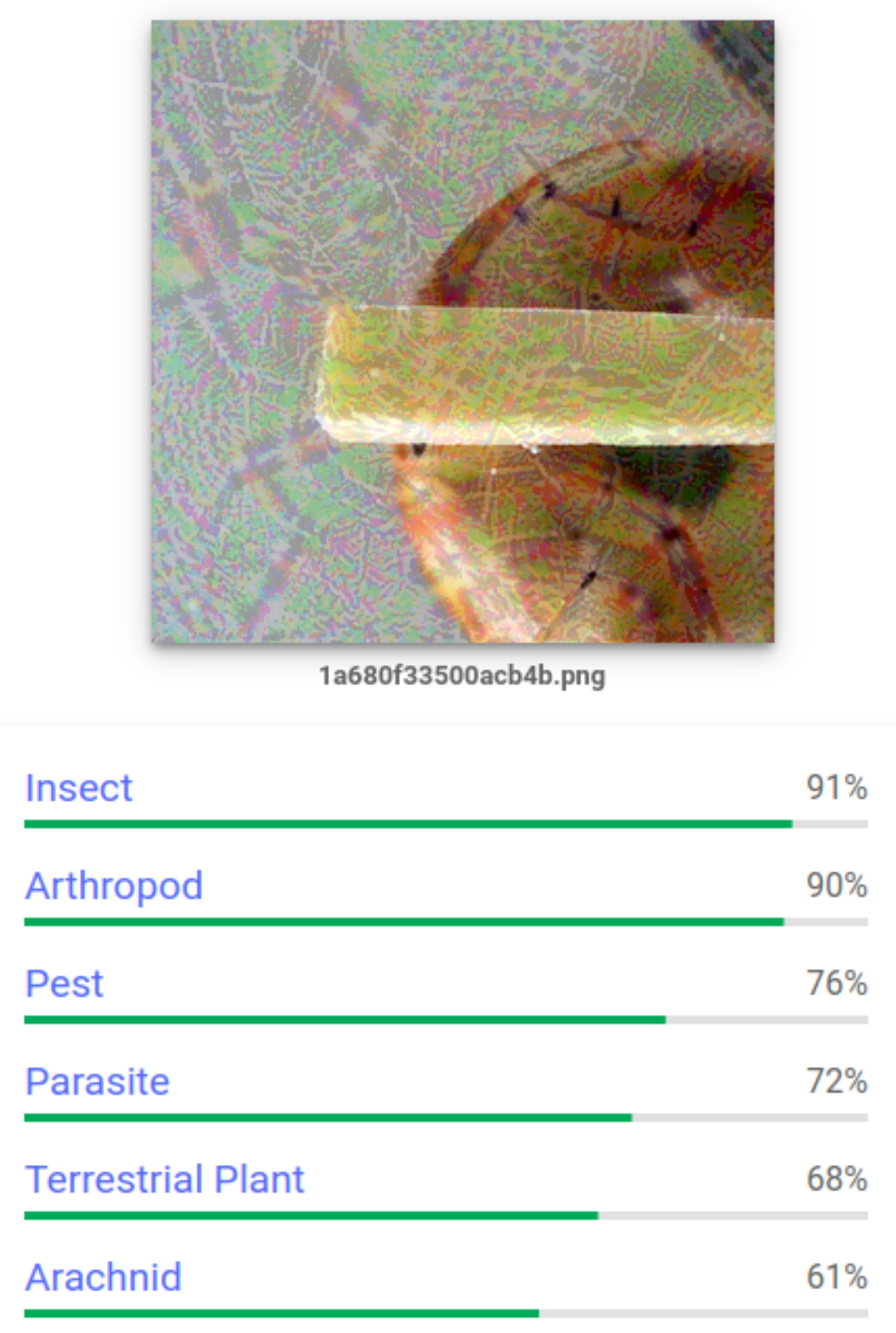} & 
  \includegraphics[width=0.18\textwidth]{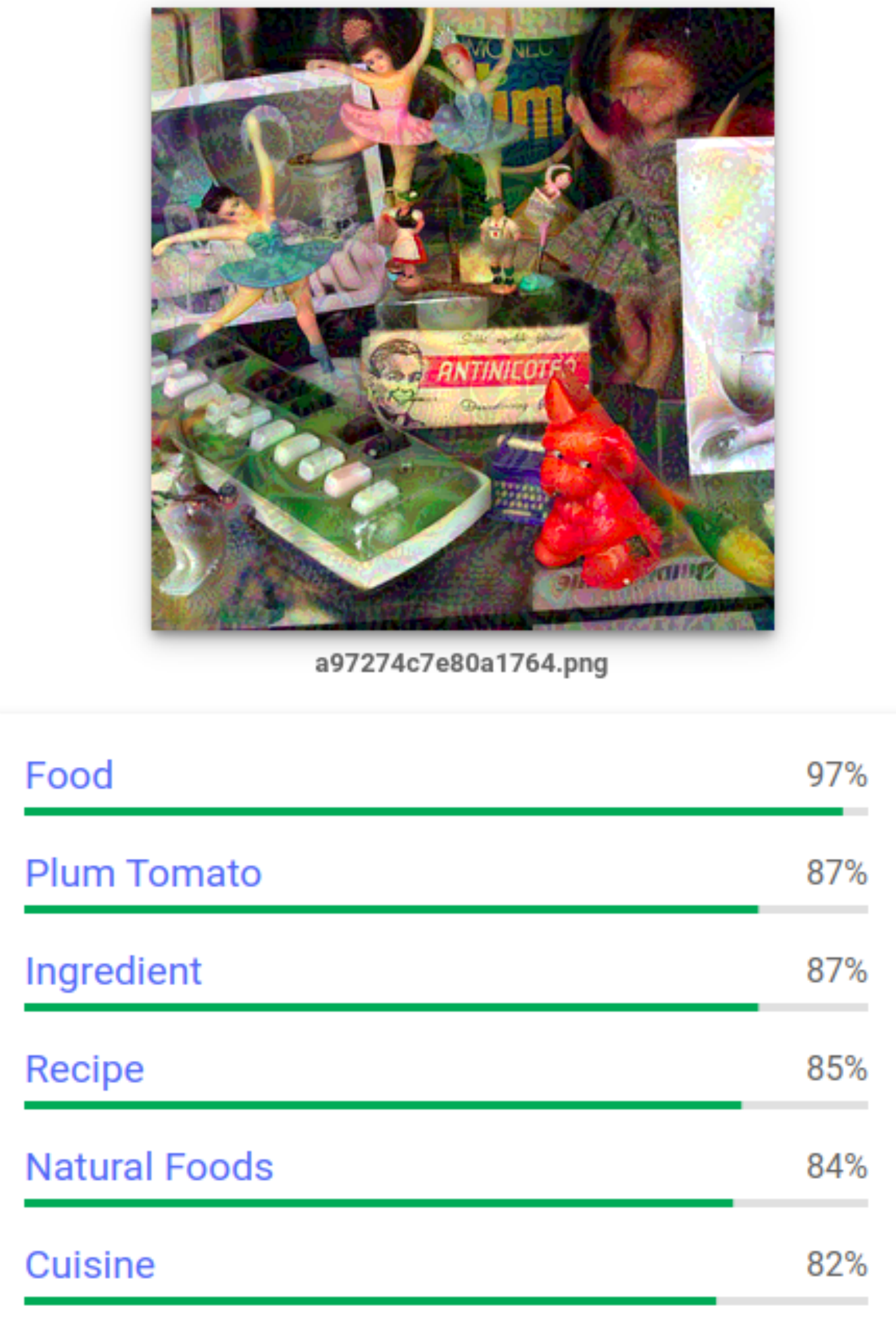} &
  \includegraphics[width=0.18\textwidth]{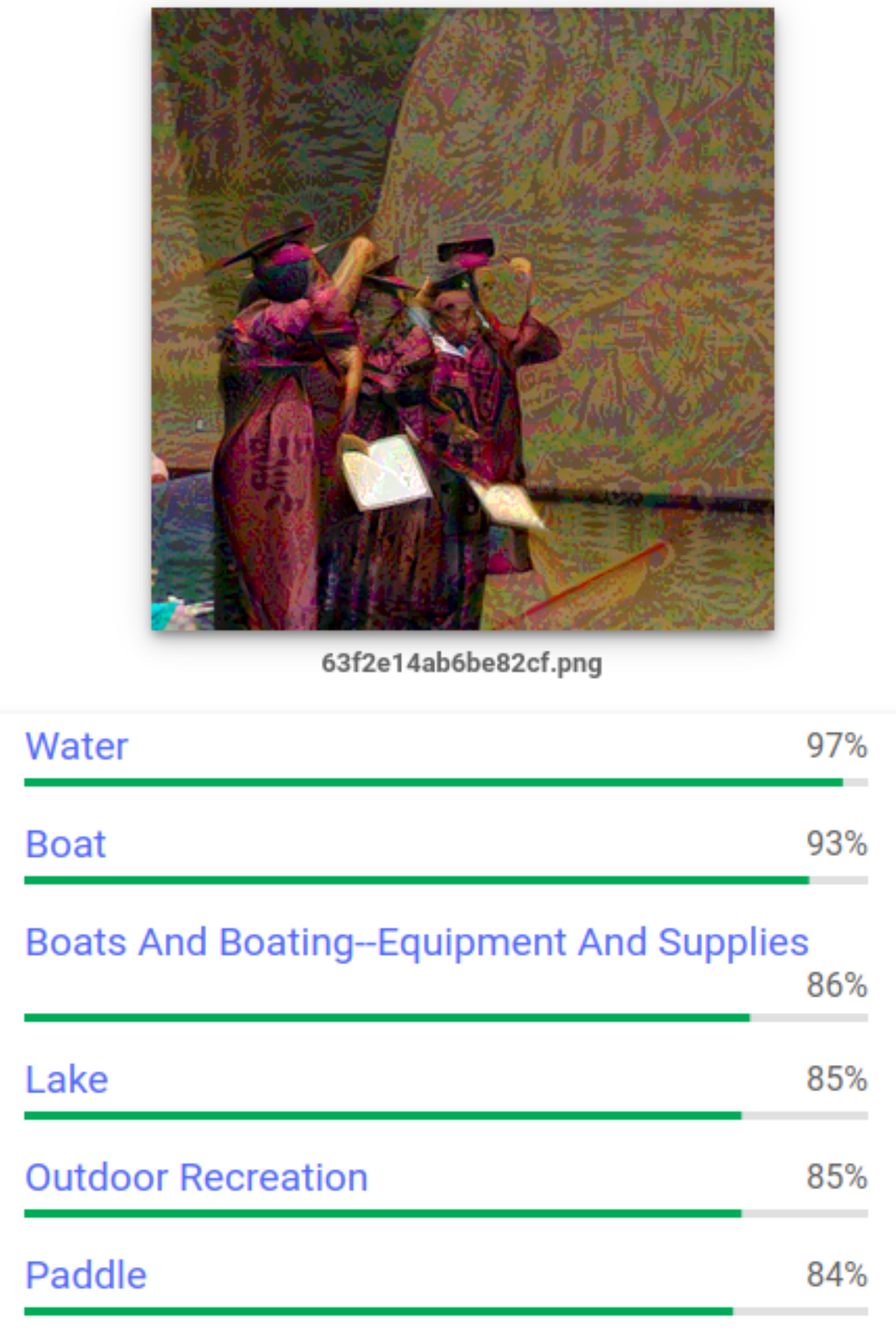} &
  \includegraphics[width=0.18\textwidth]{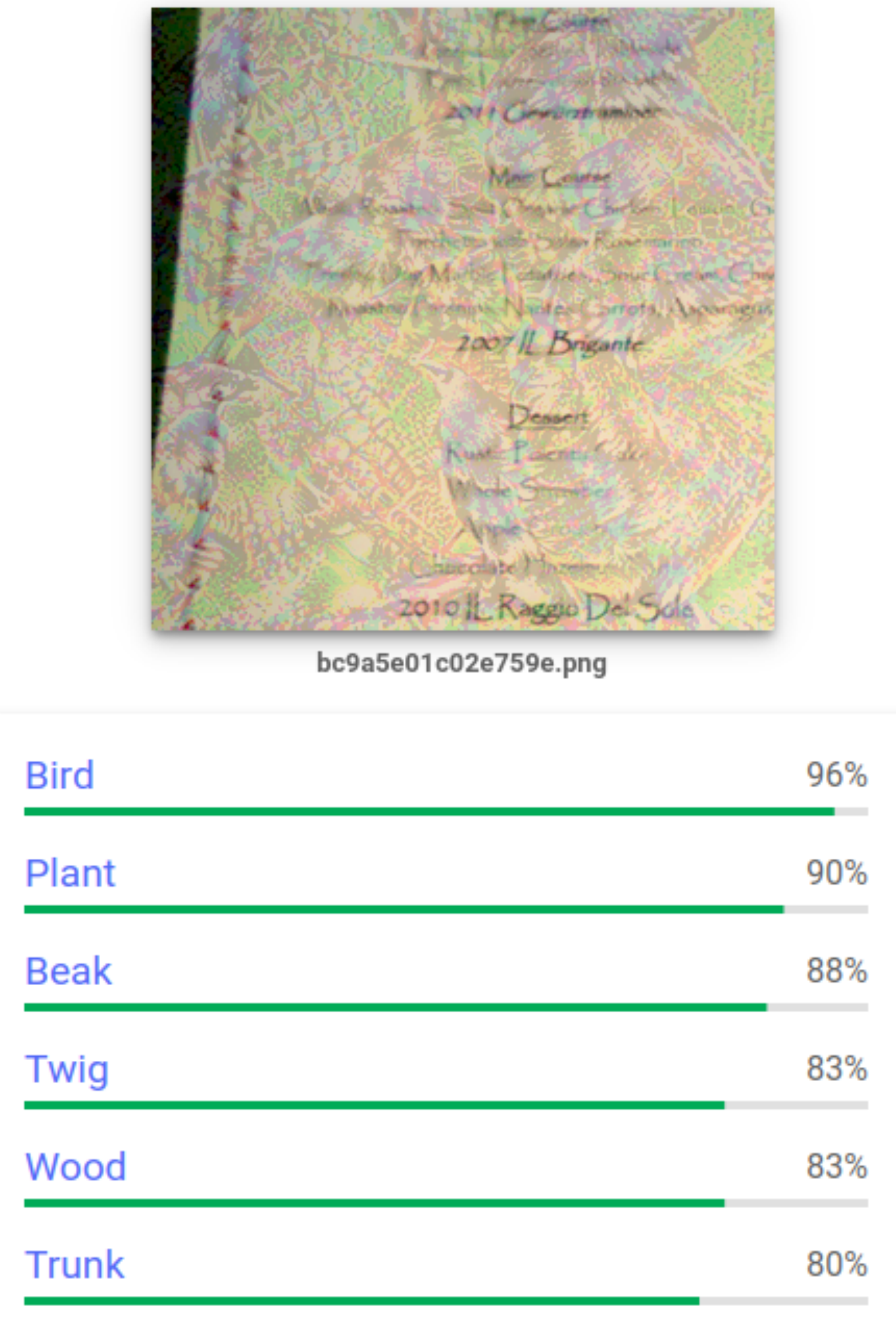} &
  \includegraphics[width=0.18\textwidth]{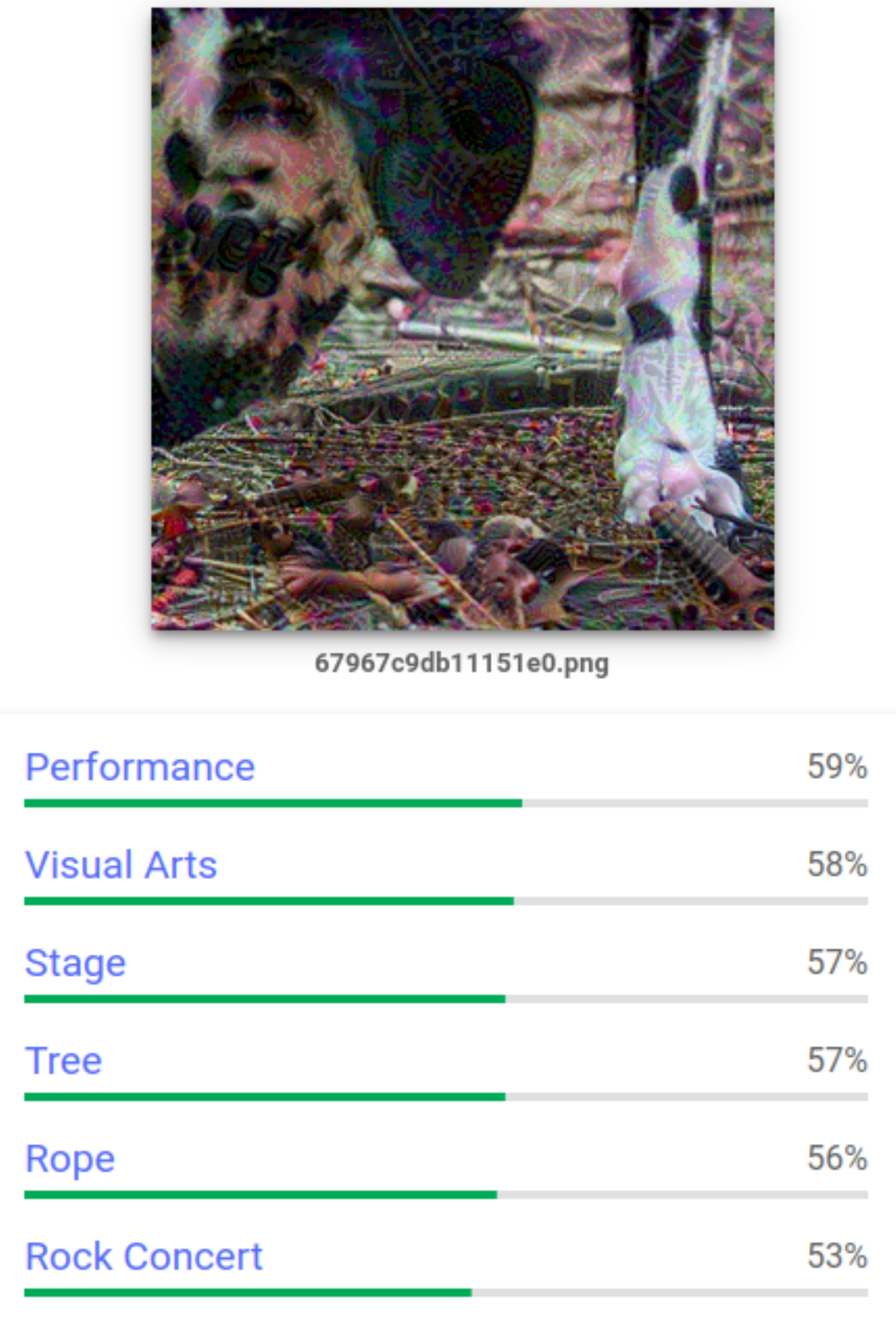}\\
Bagel $\rightarrow$ Spider & Toy Shop $\rightarrow$ Consomme & Mortarboard $\rightarrow$ Paddle & Menu $\rightarrow$ Jay  & Dog $\rightarrow$ Stage \\[6pt]

  \includegraphics[width=0.18\textwidth]{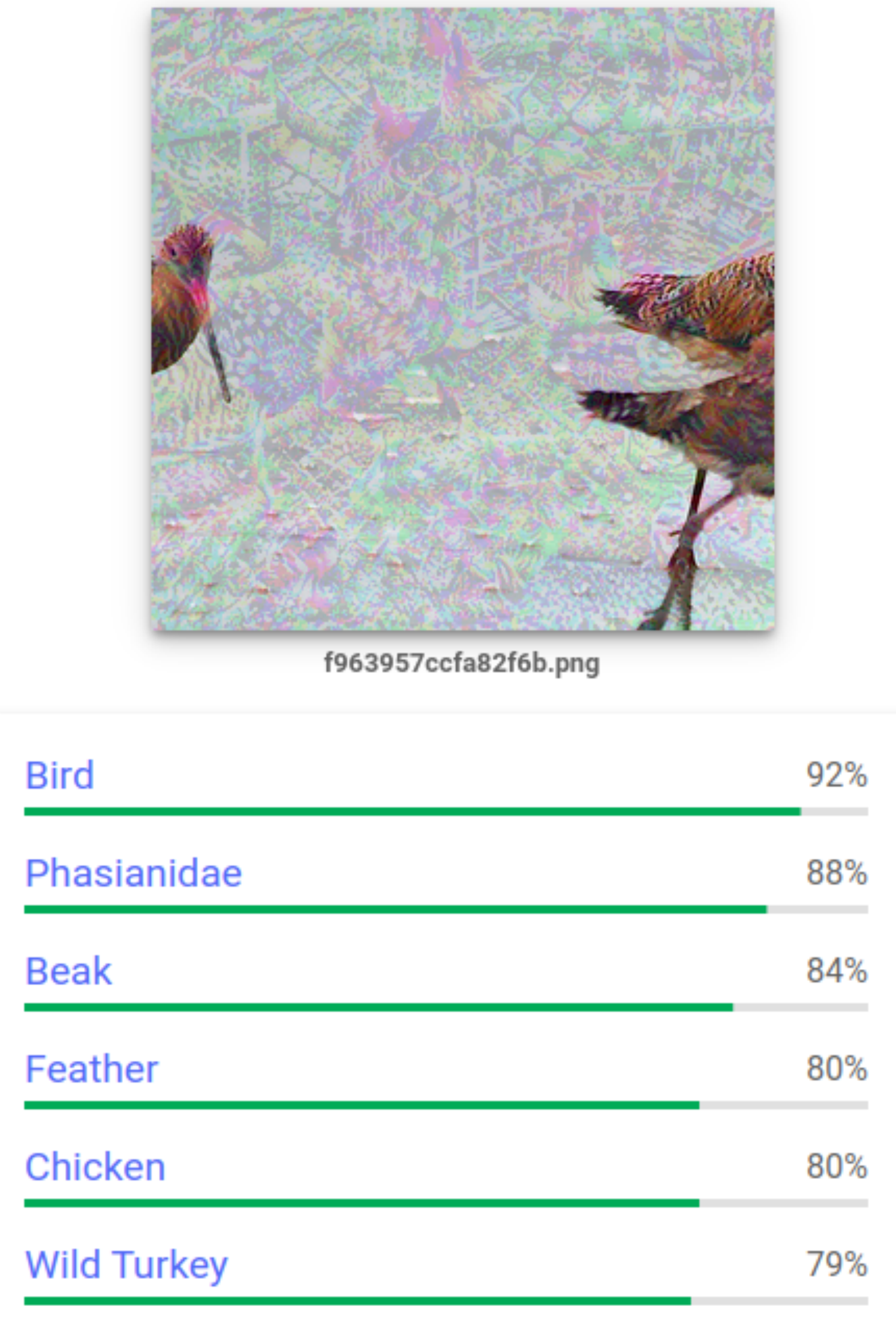} & 
  \includegraphics[width=0.18\textwidth]{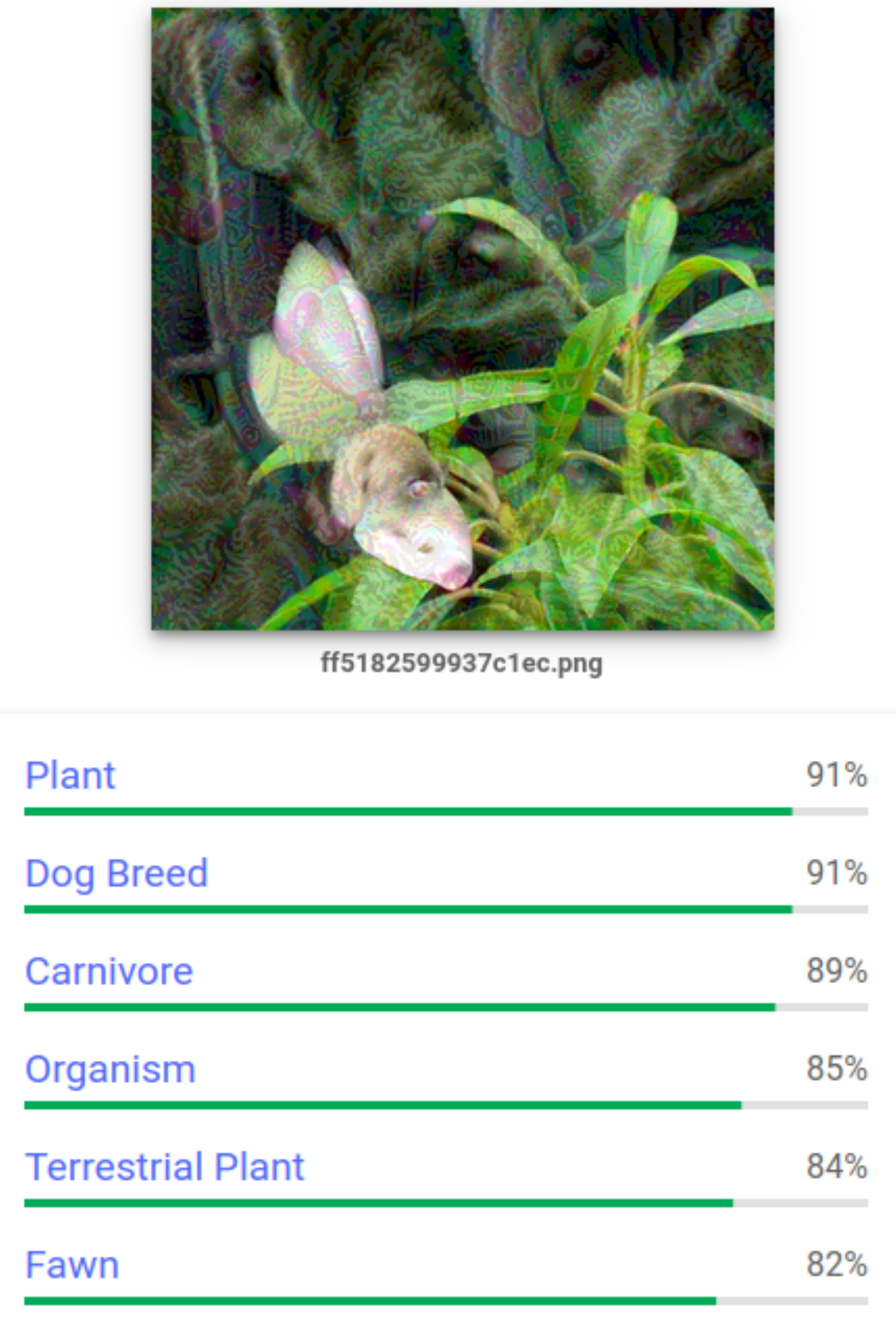} &
  \includegraphics[width=0.18\textwidth]{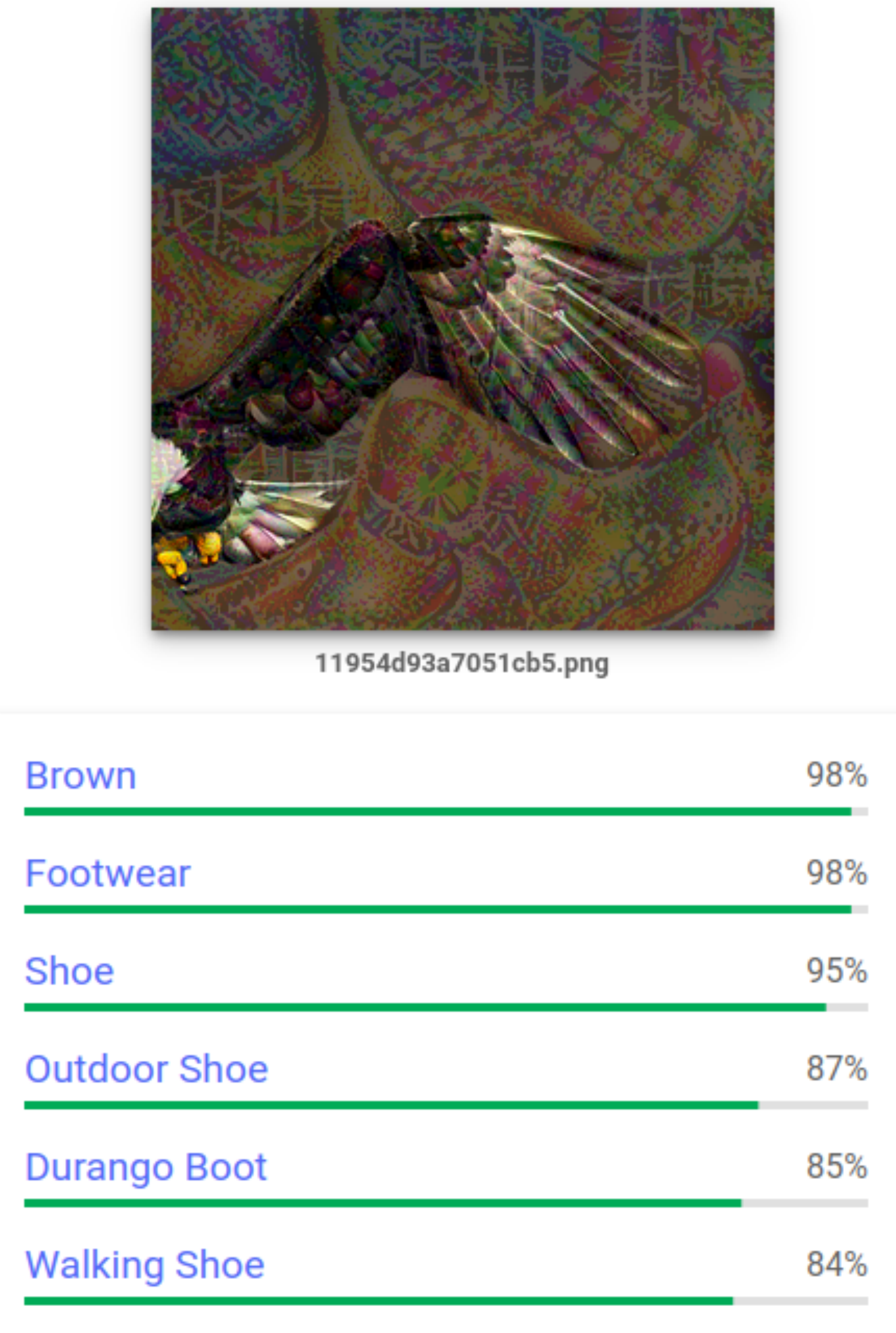} &
  \includegraphics[width=0.18\textwidth]{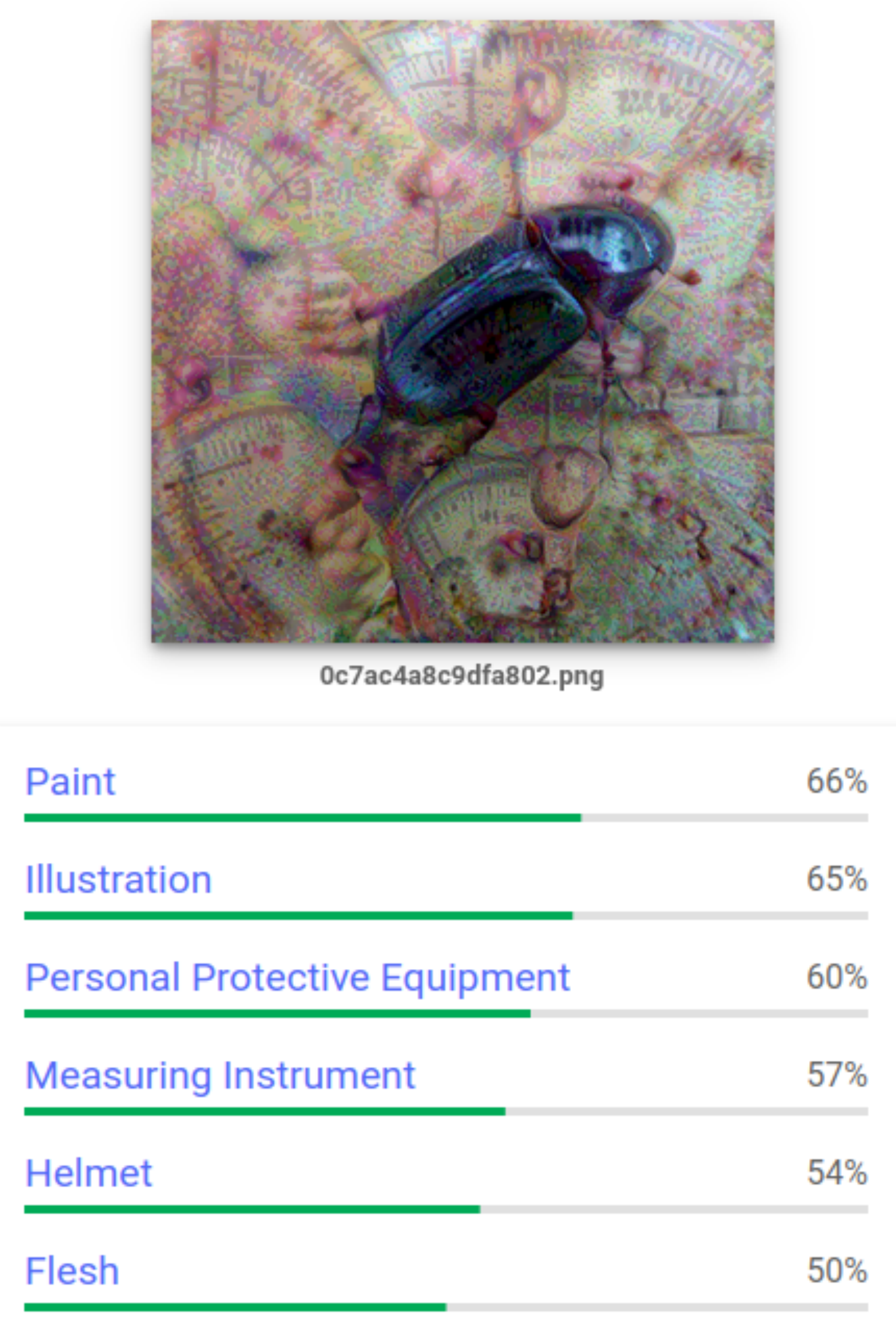} &
  \includegraphics[width=0.18\textwidth]{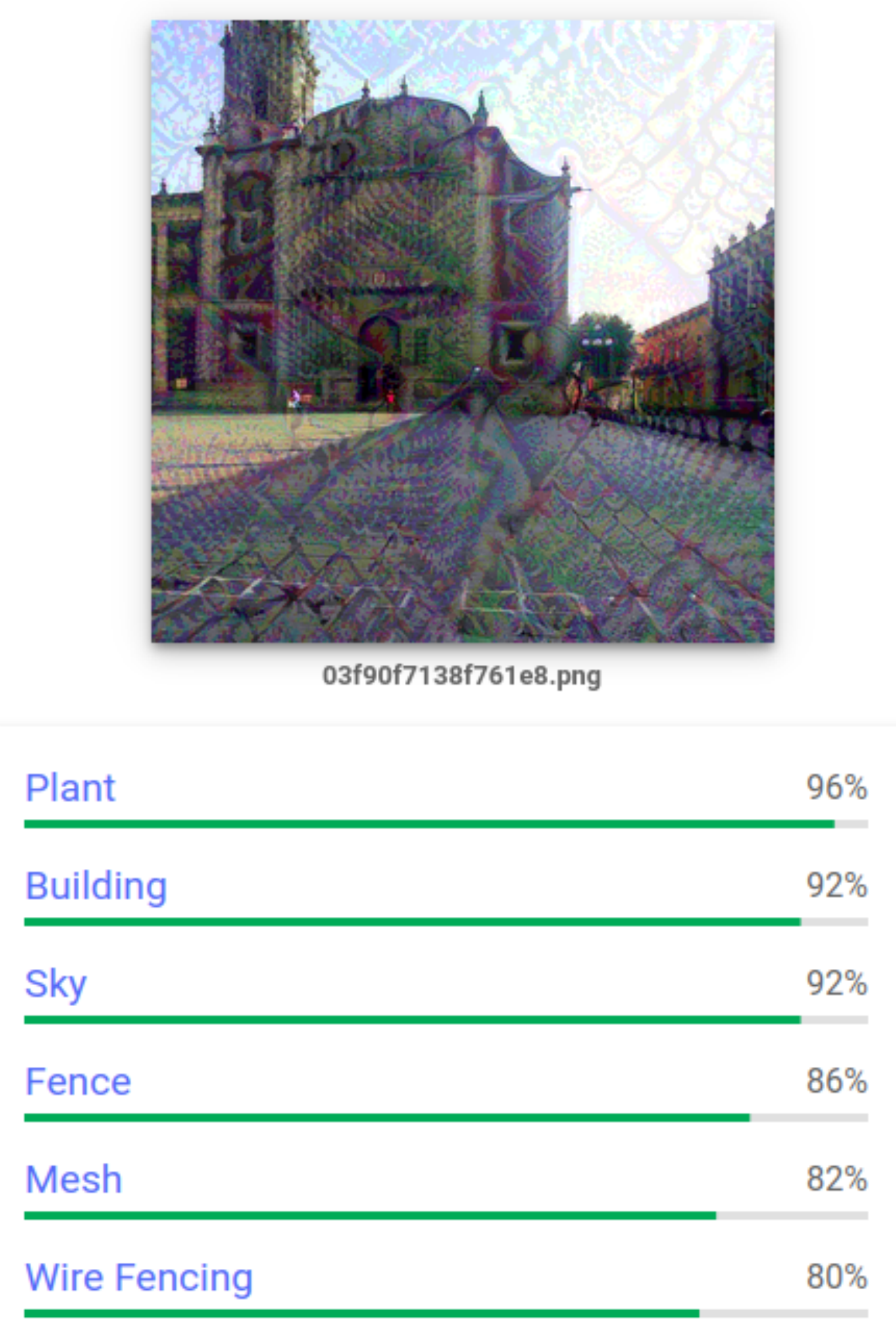}\\
Dowitcher $\rightarrow$ Cock & Butterfly $\rightarrow$ Dog  & Eagle $\rightarrow$ Geta & Beetle $\rightarrow$ Weight Machine  & Monastery $\rightarrow$ Fence \\[6pt]

  \includegraphics[width=0.18\textwidth]{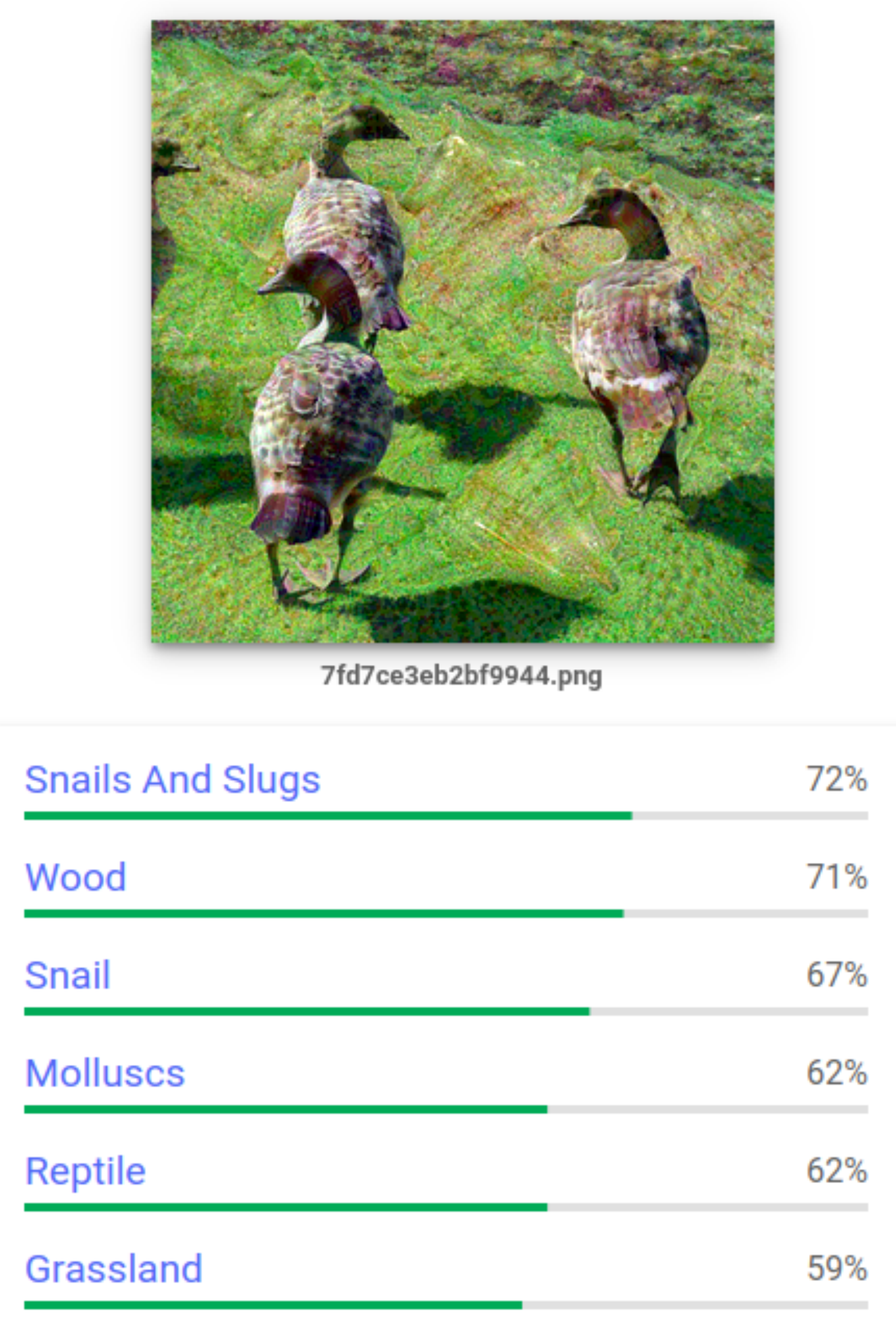} & 
  \includegraphics[width=0.18\textwidth]{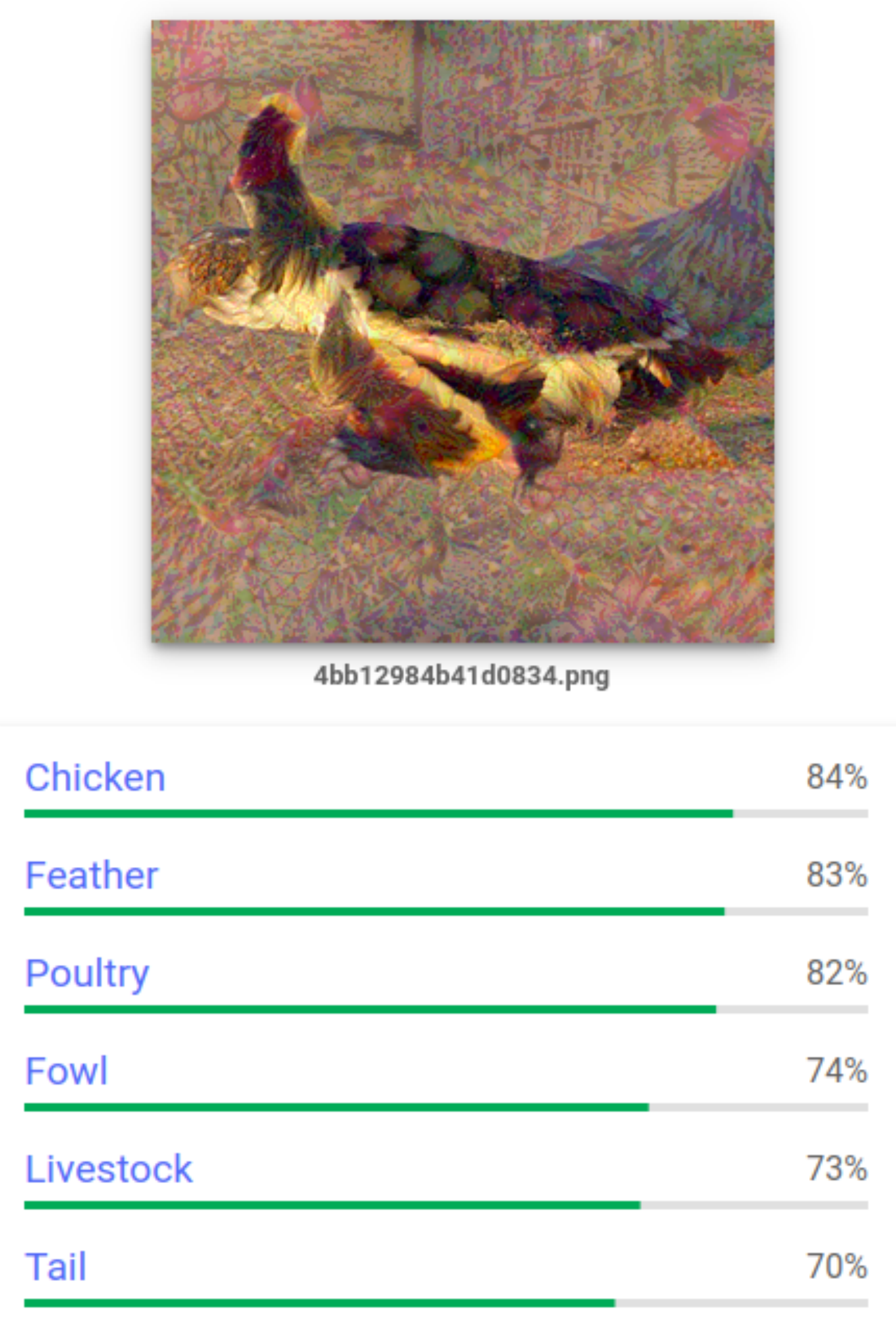} &
  \includegraphics[width=0.18\textwidth]{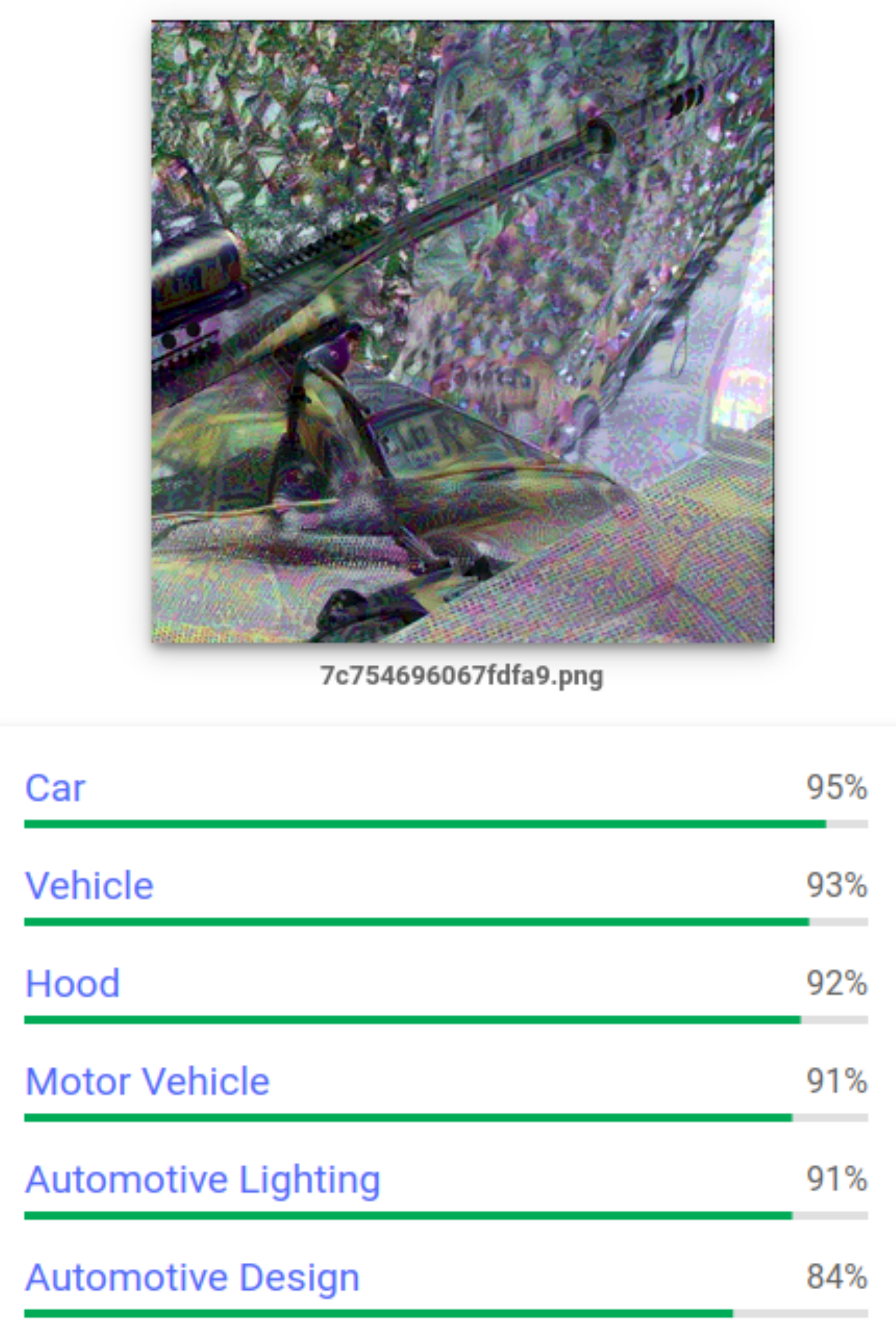} &
  \includegraphics[width=0.18\textwidth]{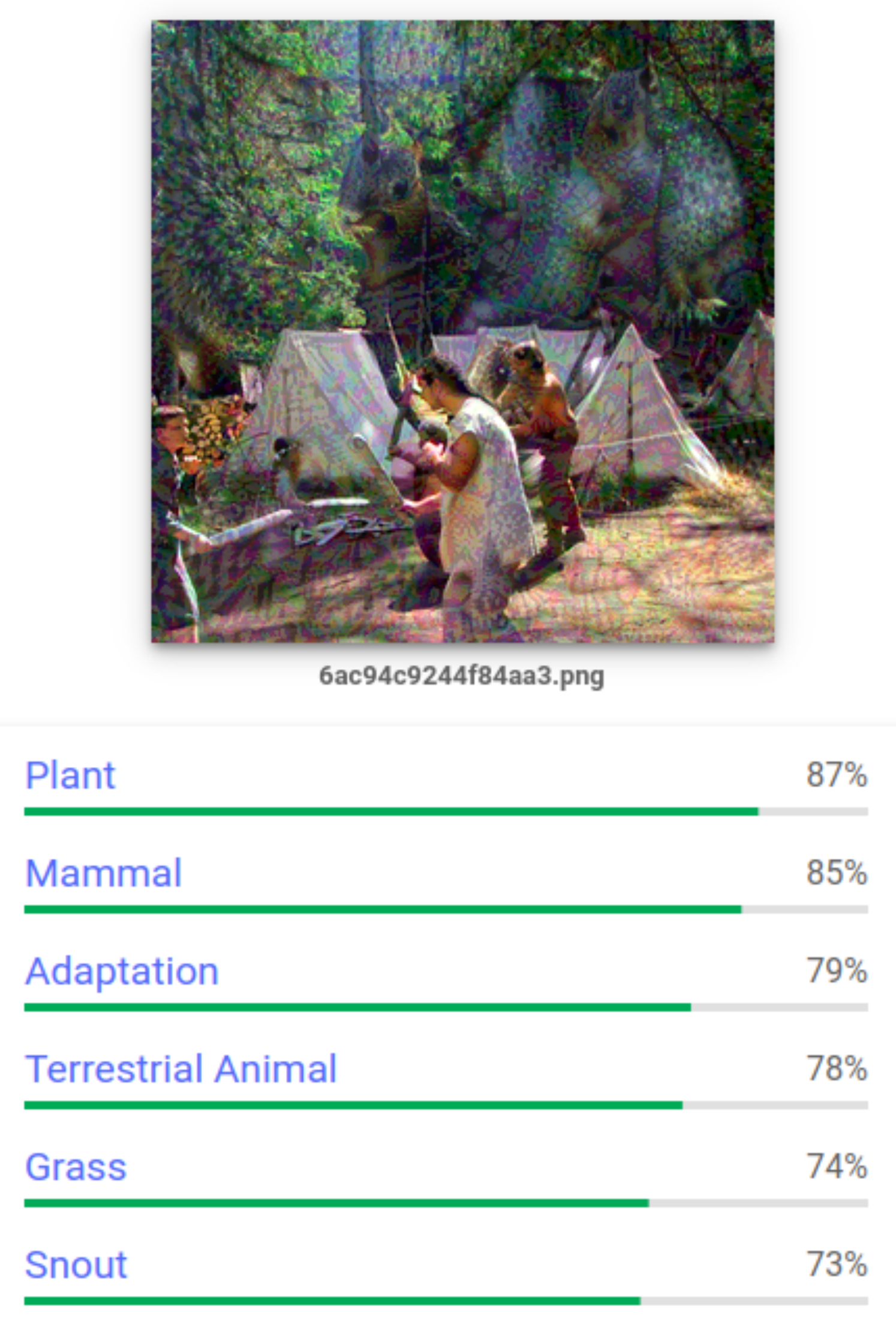} &
  \includegraphics[width=0.18\textwidth]{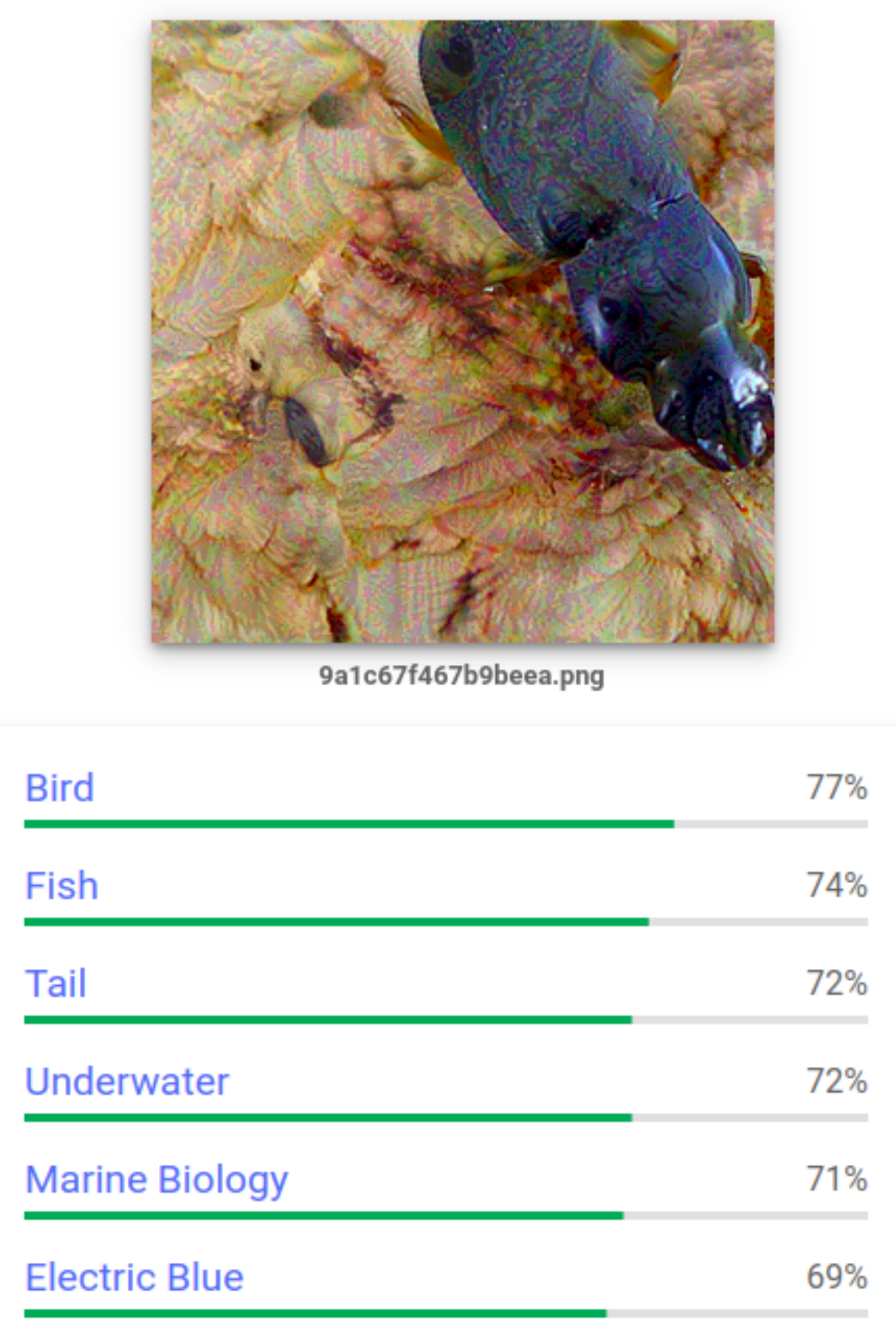}\\
Goose $\rightarrow$ Conch & Turtle $\rightarrow$ Cock & Rifle $\rightarrow$ Taxi & Fox $\rightarrow$ Squirrel & Beetle $\rightarrow$ Cockatoo \\[6pt]

  \includegraphics[width=0.18\textwidth]{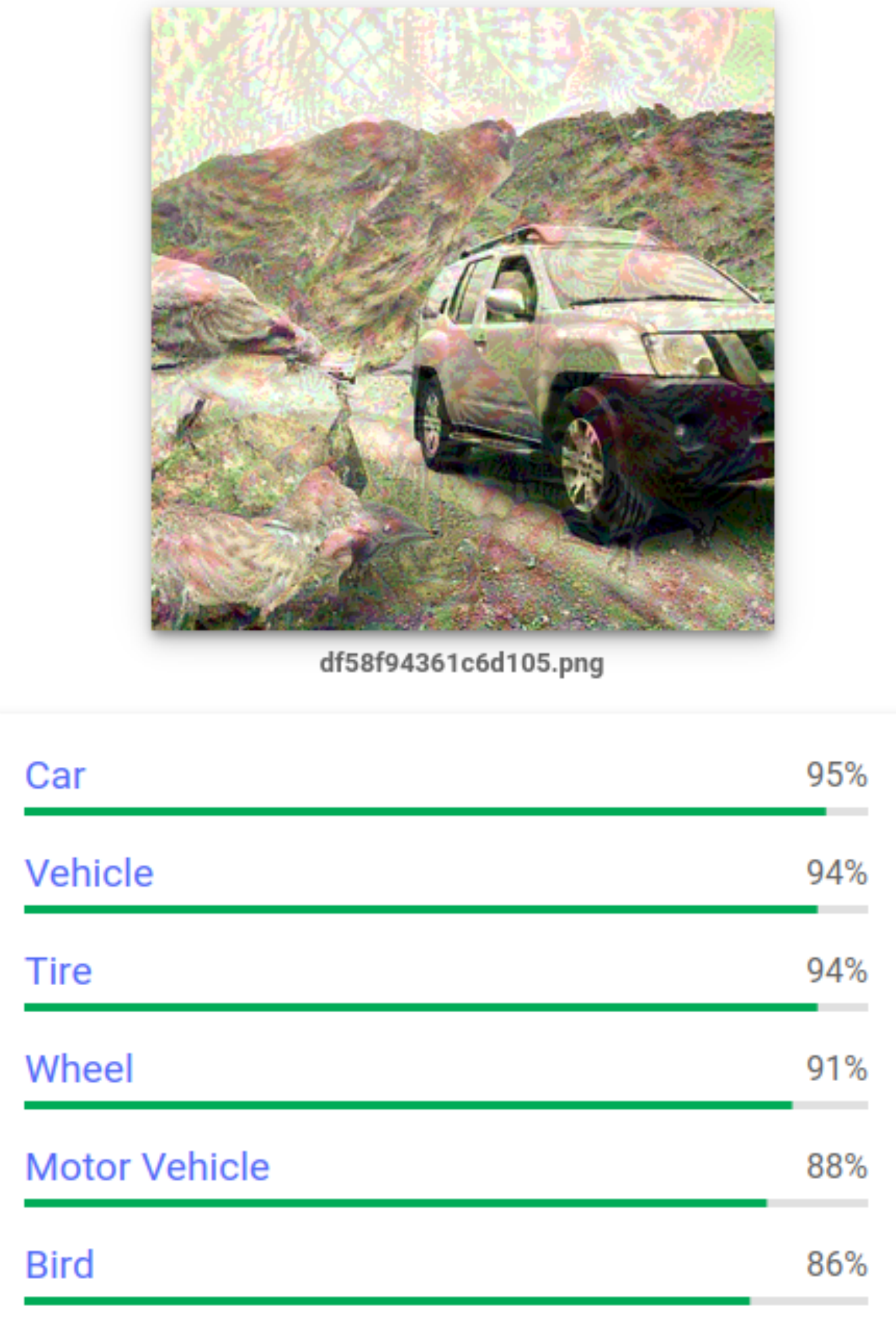} & 
  \includegraphics[width=0.18\textwidth]{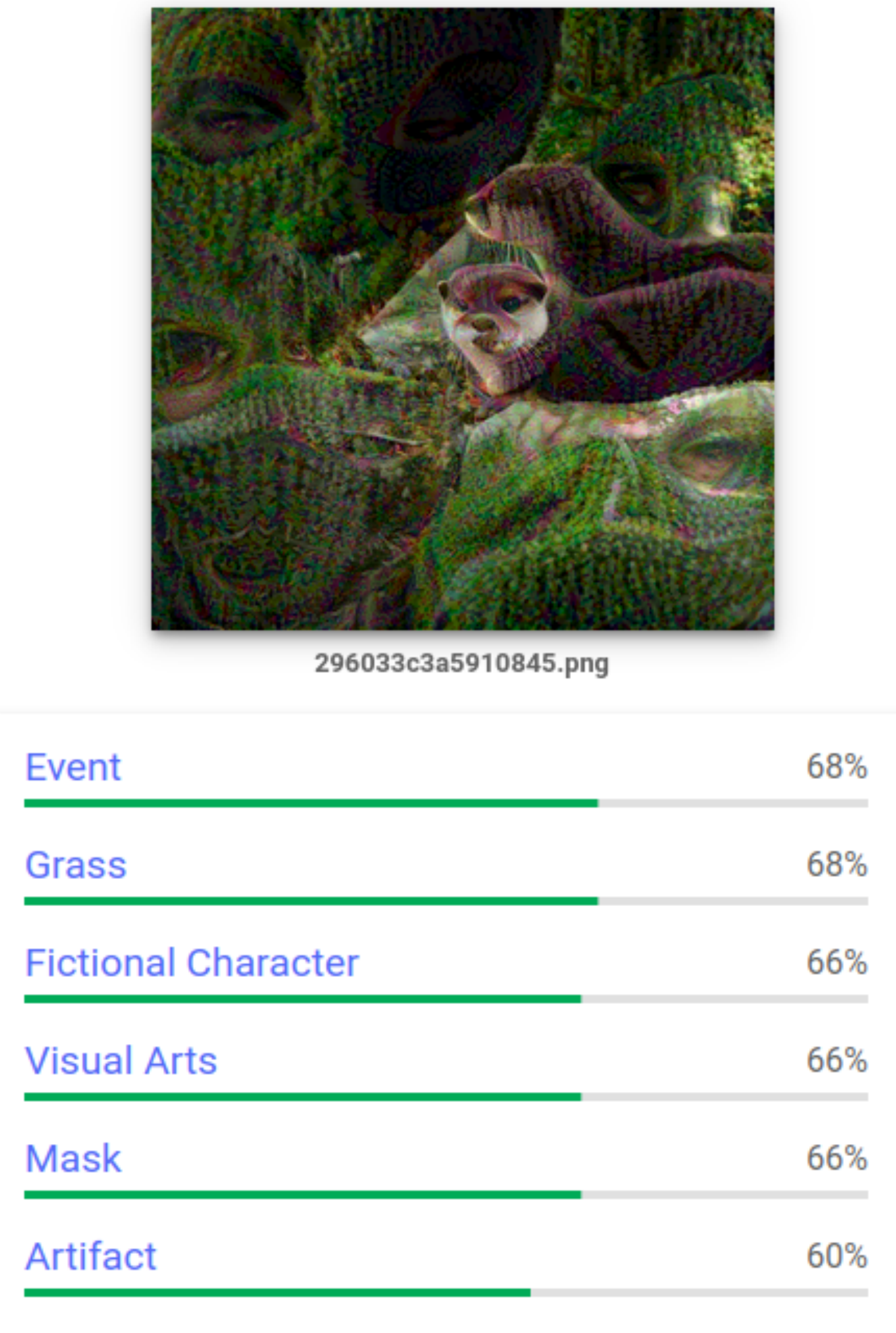} &
  \includegraphics[width=0.18\textwidth]{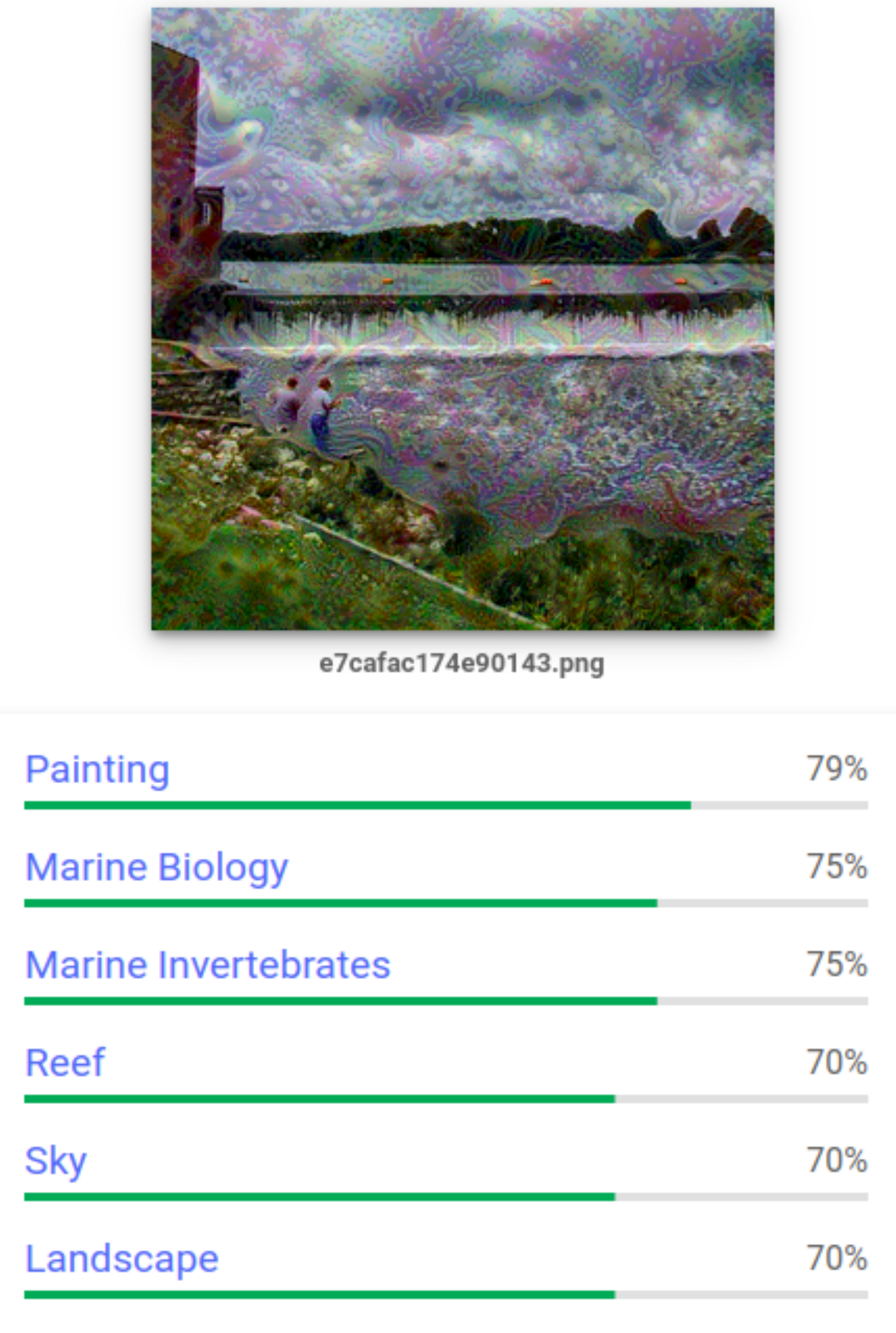} &
  \includegraphics[width=0.18\textwidth]{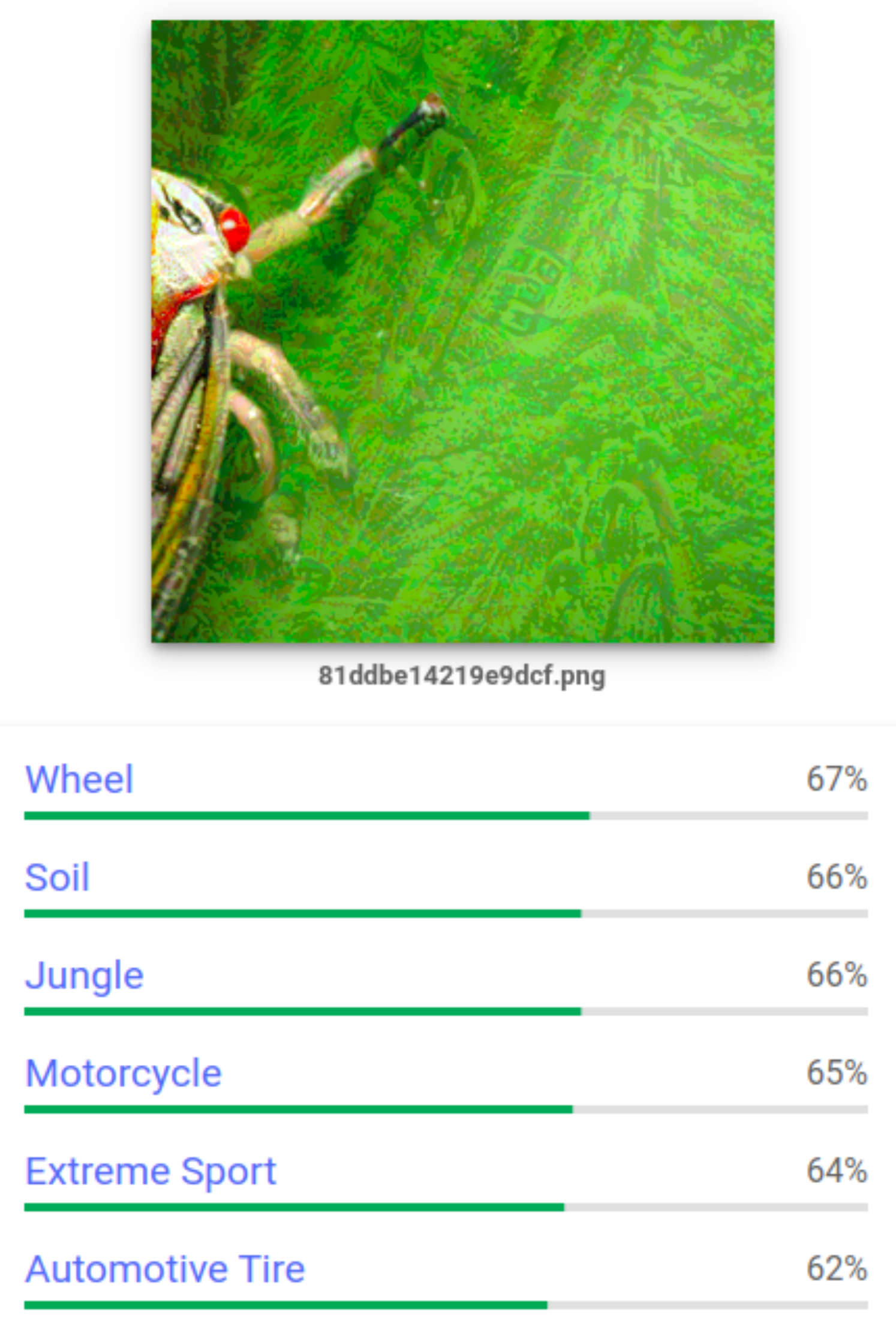} &
  \includegraphics[width=0.18\textwidth]{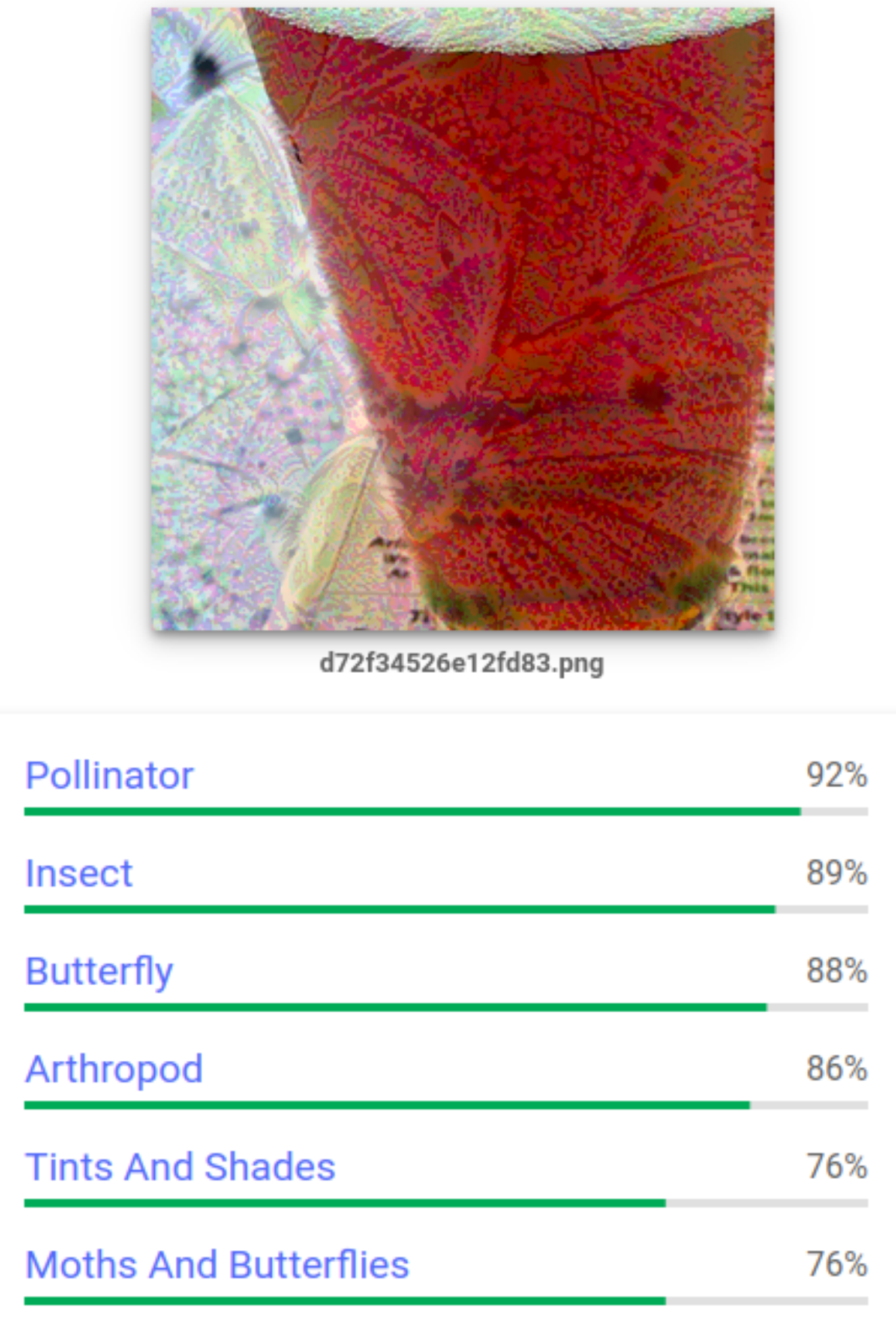}\\
Jeep $\rightarrow$ Linnet & Otter $\rightarrow$ Mask & Dam $\rightarrow$ Sea Slug & Leaf Hopper $\rightarrow$ Bike & Beer Glass $\rightarrow$ Butterfly \\[6pt]

\end{tabular}

\newpage
\clearpage

\section{Samples of GradCAM}

\addtolength{\tabcolsep}{-2pt}
\hspace{-9mm}
\begin{tabular}{cccccc}
    \includegraphics[width=0.15\textwidth]{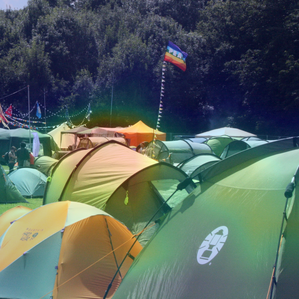} &
    \includegraphics[width=0.15\textwidth]{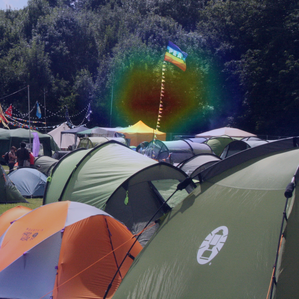} &
    \includegraphics[width=0.15\textwidth]{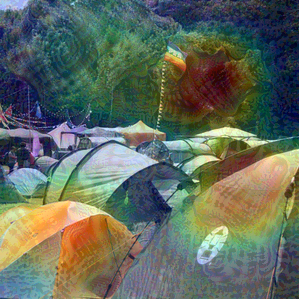} & 
    \includegraphics[width=0.15\textwidth]{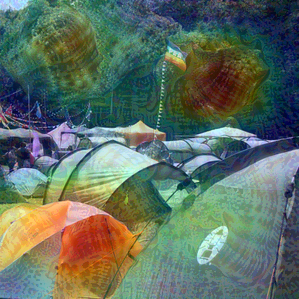} &
    \includegraphics[width=0.15\textwidth]{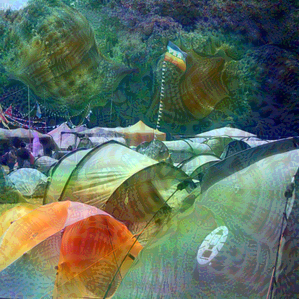}&
    \includegraphics[width=0.15\textwidth]{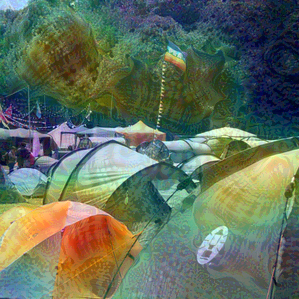}\\
    Clean \textbf{Tent} &
    Clean \textbf{Conch} &
    NI-SI-TI-DI \textbf{Conch} &
    +\ghost{} \textbf{Conch} &
    +\dual{} \textbf{Conch} &
    +DWP \textbf{Conch} \\[6pt]

    \includegraphics[width=0.15\textwidth]{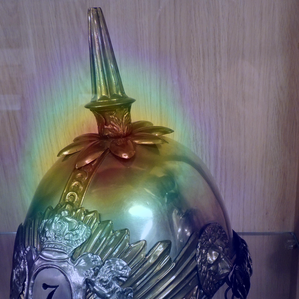} &
    \includegraphics[width=0.15\textwidth]{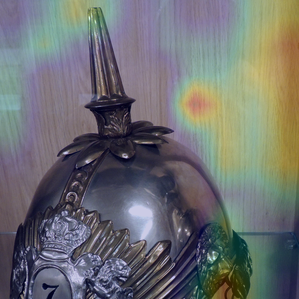} &
    \includegraphics[width=0.15\textwidth]{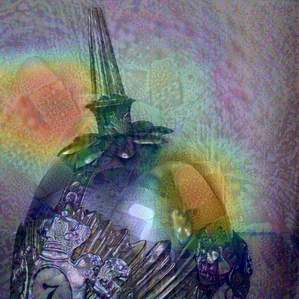} & 
    \includegraphics[width=0.15\textwidth]{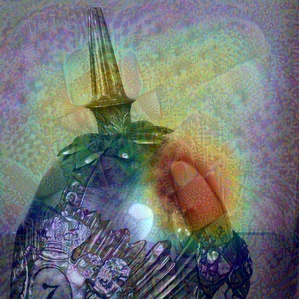} &
    \includegraphics[width=0.15\textwidth]{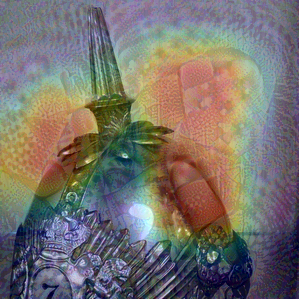}&
    \includegraphics[width=0.15\textwidth]{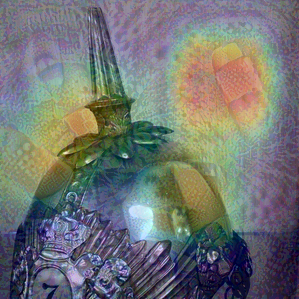}\\
    Clean \textbf{Pickelhaube} &
    Clean \textbf{Band Aid} &
    NI-SI-TI-DI \textbf{Band Aid} &
    +\ghost{} \textbf{Band Aid} &
    +\dual{} \textbf{Band Aid} &
    +DWP \textbf{Band Aid} \\[6pt]

    \includegraphics[width=0.15\textwidth]{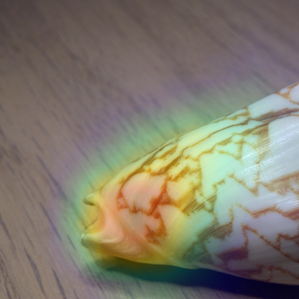} &
    \includegraphics[width=0.15\textwidth]{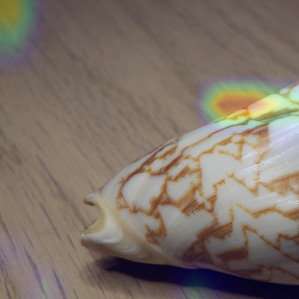} &
    \includegraphics[width=0.15\textwidth]{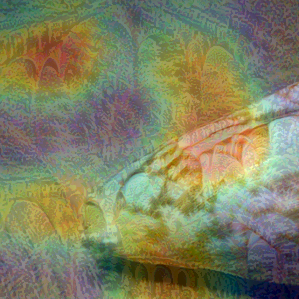} & 
    \includegraphics[width=0.15\textwidth]{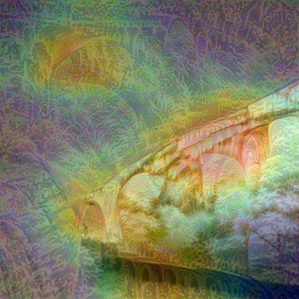} &
    \includegraphics[width=0.15\textwidth]{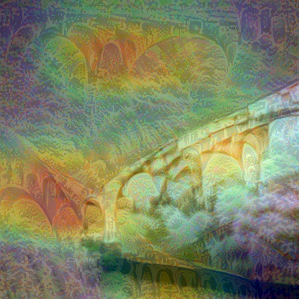}&
    \includegraphics[width=0.15\textwidth]{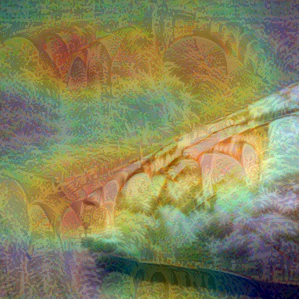}\\
    Clean \textbf{Conch} &
    Clean \textbf{Viaduct} &
    NI-SI-TI-DI \textbf{Viaduct} &
    +\ghost{} \textbf{Viaduct} &
    +\dual{} \textbf{Viaduct} &
    +DWP \textbf{Viaduct} \\[6pt]

    \includegraphics[width=0.15\textwidth]{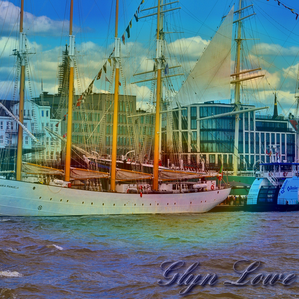} &
    \includegraphics[width=0.15\textwidth]{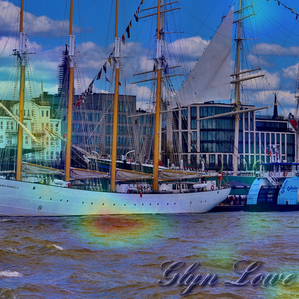} &
    \includegraphics[width=0.15\textwidth]{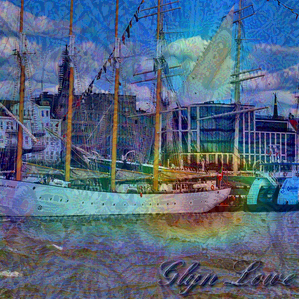} & 
    \includegraphics[width=0.15\textwidth]{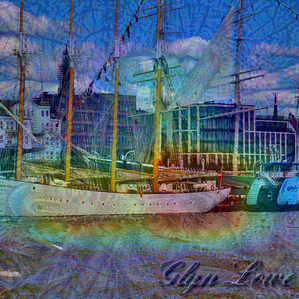} &
    \includegraphics[width=0.15\textwidth]{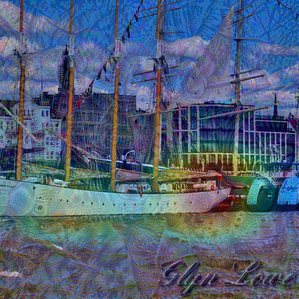}&
    \includegraphics[width=0.15\textwidth]{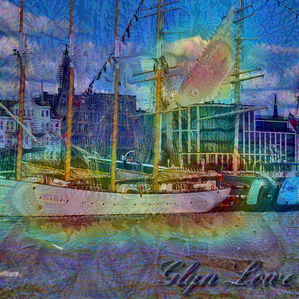}\\
    Clean \textbf{Schooner} &
    Clean \textbf{Chainsaw} &
    NI-SI-TI-DI \textbf{Chainsaw} &
    +\ghost{} \textbf{Chainsaw} &
    +\dual{} \textbf{Chainsaw} &
    +DWP \textbf{Chainsaw} \\[6pt]

    \includegraphics[width=0.15\textwidth]{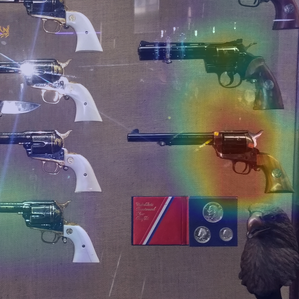} &
    \includegraphics[width=0.15\textwidth]{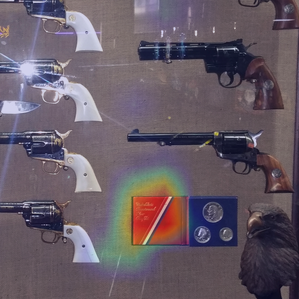} &
    \includegraphics[width=0.15\textwidth]{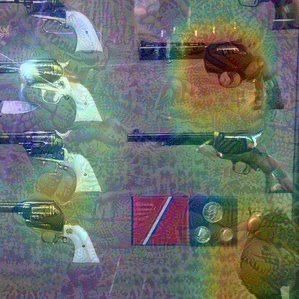} & 
    \includegraphics[width=0.15\textwidth]{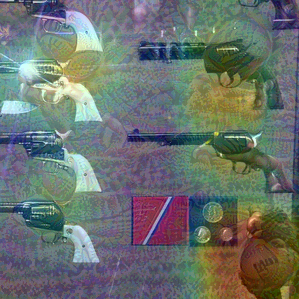} &
    \includegraphics[width=0.15\textwidth]{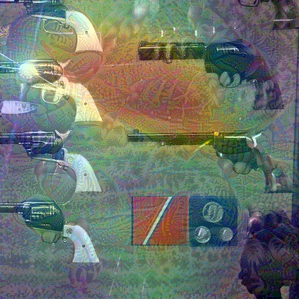}&
    \includegraphics[width=0.15\textwidth]{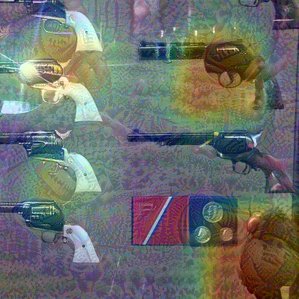}\\
    Clean \textbf{Revolver} &
    Clean \textbf{Rugby Ball} &
    NI-SI-TI-DI \textbf{Rugby Ball} &
    +\ghost{} \textbf{Rugby Ball} &
    +\dual{} \textbf{Rugby Ball} &
    +DWP \textbf{Rugby Ball} \\[6pt]

    \includegraphics[width=0.15\textwidth]{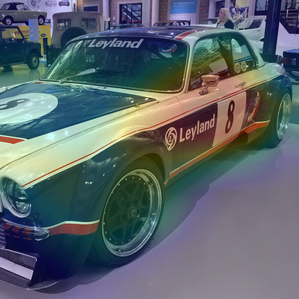} &
    \includegraphics[width=0.15\textwidth]{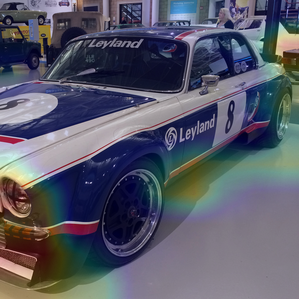} &
    \includegraphics[width=0.15\textwidth]{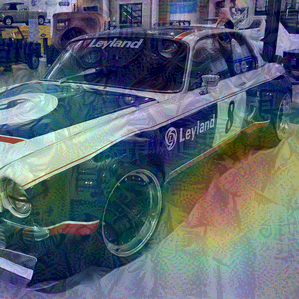} & 
    \includegraphics[width=0.15\textwidth]{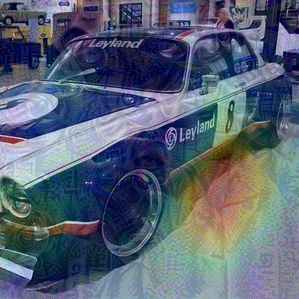} &
    \includegraphics[width=0.15\textwidth]{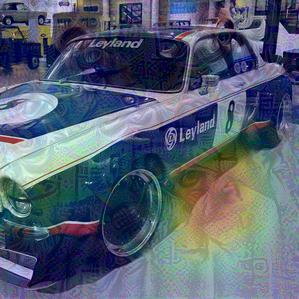}&
    \includegraphics[width=0.15\textwidth]{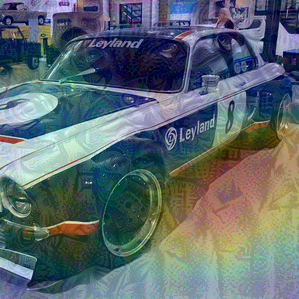}\\
    Clean \textbf{Racing Car} &
    Clean \textbf{Lab Coat} &
    NI-SI-TI-DI \textbf{Lab Coat} &
    +\ghost{} \textbf{Lab Coat} &
    +\dual{} \textbf{Lab Coat} &
    +DWP \textbf{Lab Coat} \\[6pt]

\end{tabular}
\addtolength{\tabcolsep}{2pt}

\newpage
\clearpage

\end{document}